\documentclass{article}

\usepackage[preprint]{Styles/neurips_2024}

\usepackage[utf8]{inputenc} 
\usepackage[T1]{fontenc}    
\usepackage{hyperref}       
\usepackage{url}            
\usepackage{booktabs}       
\usepackage{amsfonts}       
\usepackage{nicefrac}       
\usepackage{microtype}      
\usepackage{xcolor}         
\usepackage{multirow}
\usepackage{booktabs}
\usepackage{subcaption}
\usepackage{amsmath}
\usepackage{graphicx}
\usepackage{wrapfig}
\def\clas{\textsl{CLAS}}
\def\shapenet{\textsc{ShapeNet}}
\title{\clas{}: A Machine Learning Enhanced Framework for Exploring Large 3D Design Datasets}

\author{XiuYu Zhang\thanks{Communication to xiuyuzhang@berkeley.edu} \\
  EECS, UC Berkeley \\ \\\And
  Xiaolei Ye \\
  ME, UC Berkeley \\ \\\And
  Jui-Che Chang \\
  ME, UC Berkeley \\ \\\And
  Yue Fang \\
  ME, UC Berkeley \\ \\}

\begin{document}

\maketitle

\begin{abstract}
  Three-dimensional (3D) objects have wide applications. Despite the growing interest in 3D modeling in academia and industries, designing and/or creating 3D objects from scratch remains time-consuming and challenging. With the development of generative artificial intelligence (AI), designers discover a new way to create images for ideation. However, generative AIs are less useful in creating 3D objects with satisfying qualities. To allow 3D designers to access a wide range of 3D objects for creative activities based on their specific demands, we propose a machine learning (ML) enhanced framework \clas{} - named after the four-step of capture, label, associate, and search - to enable fully automatic retrieval of 3D objects based on user specifications leveraging the existing datasets of 3D objects. \clas{} provides an effective and efficient method for any person or organization to benefit from their existing but not utilized 3D datasets. In addition, \clas{} may also be used to produce high-quality 3D object synthesis datasets for training and evaluating 3D generative models. As a proof of concept, we created and showcased a search system with a web user interface (UI) for retrieving 6,778 3D objects of chairs in \textsc{ShapeNet} dataset powered by \clas{}. In a close-set retrieval setting, our retrieval method achieves a mean reciprocal rank (MRR) of 0.58, top 1 accuracy of 42.27\%, and top 10 accuracy of 89.64\%. 
\end{abstract}

\section{Introduction}\label{sec:introduction}

The creative design process has been transformed by the advance of generative artificial intelligence (AI), benefiting designers in different fields~\citep{picard2023concept}. Text-to-text large language models (LLM) such as OpenAI's \textsc{ChatGPT}~\citep{chatgpt} and Meta's ~\textsc{LLaMA}\citep{touvron2023llama, Touvron2023Llama2O} provide a convenient and powerful way for designers to formulate their design ideas~\citep{Zhu2021GenerativePT}. In addition to discussing ideas in words, text-to-image generative AIs could help designers visualize and showcase their thoughts much faster. The wide adoption of latent diffusion models~\citep{rombach2022highresolution} demonstrated by the popularity of \textsc{Stable Diffusion} and \textsc{Midjourney} proves that the use of text-to-image models is welcomed by many designers. However, the experience of designers immersed in 3D modeling might be different. It is natural and necessary to extend the success of AI-generated content (AIGC) to the three-dimensional (3D) domain, as 3D objects have many applications in diverse fields~\citep{li2023generative, NEURIPS2021_31445061,10.1007/978-3-031-25066-8_39}. However, due to the additional dimension of the output data and the relatively restricted datasets for training, the performance of the state-of-art 3D generative models developed on Generative adversarial network (GAN)~\citep{Goodfellow2014GenerativeAN, Chan2021EfficientG3, 3dgp}, Neural radiance fields (Nerf)~\citep{mildenhall2020nerf, NEURIPS2023_525d2440}, 3D diffusion models~\citep{liu2023one2345++, Magic123} or hybrid approaches, is not on par with what has been offered by text-to-text or text-to-image models. Generated 3D objects may have lower visual quality, distorted geometry, and not fully complete 360-degree generation~\citep{3dgp}. In addition, 3D object generations perform better when intensive information is given as prior, such as multi-view images and 3D objects~\citep{liu2023one2345++, Magic123}. Since in the ideation phase of design, 3D designers are looking for references and inspirations, such detailed prior can hardly be provided, which limits the use of the current limited 3D generation. The training data of text and image generation models can be constructed from large quantities of diverse sources since these modalities can be created and accessed by the general public. In contrast, 3D objects are generally created by professionals for professional tasks, resulting in many 3D objects being proprietary properties. 

\begin{figure}[t]
    \centering
    \includegraphics[width=\linewidth]{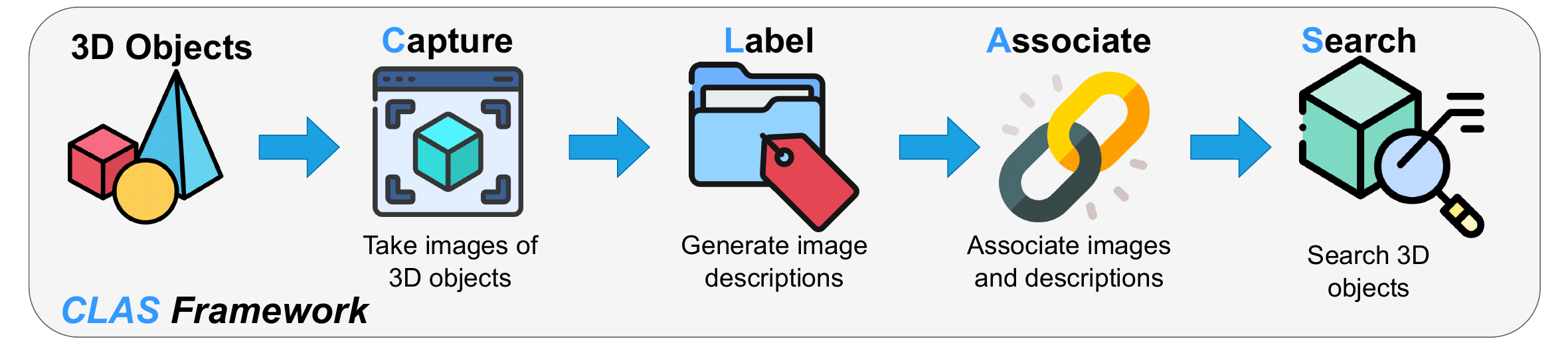}
    \caption{\clas{}, named after the main steps of capture, label, associate, and search, to enable fully automatic retrieval of 3D objects based on user specifications.}
    \label{fig:framework}
\end{figure}

Given these constraints, instead of creating a 3D object from scratch based on the designer's thoughts, providing them with relevant existing 3D objects in professional datasets might be more practical and useful. However, 3D Object Retrieval (3DOR) methods are generally based on modalities that are rich in 3D spacial information~\citep{He2018TripletCenterLF, Jing2021CrossModalCL, Feng2023HypergraphBasedMR}, such as multi-view image~\citep{Su2015MultiviewCN}, point cloud~\citep{Zhou2017VoxelNetEL} and voxel~\citep{Wu20143DSA}. These modalities are not intuitive for 3D designers to use when seeking inspiration. On the other hand, language is the most intuitive modality to describe the 3D objects designers may want as a reference. Language can be especially useful for retrievals not only based on the object's geometry but also on the varying design concepts such as use cases, intended users, and design purpose. 

We, therefore, propose a four-step framework, \textbf{\clas{} (capture, label, associate, and search)} as summarized in Figure~\ref{fig:framework}, to automatically label, index, and retrieve 3D objects based on the user's textual prompts with varying levels of details and focus. \clas{} is a flexible framework that can be applied to diverse 3D objects for different purposes. In the \textbf{capture} step, an image at a fixed position or a set of multi-view images can be created based on 3D models in the target dataset. A visual-language model is leveraged to create descriptions for these captured images based on specifically designed prompts in the \textbf{label} step. Depending on the use case, generated descriptions can have different focus (multiple descriptions can be created for an image). In the \textbf{associate} step, one or more CLIP models~\citep{radford2021learning} can be trained on the images and descriptions to learn semantic context from text and visual context from images. More than one model should be trained if more than one purpose of descriptions is generated in the previous step. Then, the CLIP models can be used in the \textbf{search} step for text-based retrieval of 3D objects. This helps designers to efficiently find inspiration, overcome creative blocks, and seamlessly initiate their design process. Furthermore, the 3D datasets labeled by \clas{} can also be repurposed to train 3D generative models with richer contextual semantic meanings. To showcase the efficiency of \clas{}, we implemented a proof of concept search system on chairs in the \textsc{ShapeNet}~\citep{shapenet2015} and evaluated its effectiveness.

\section{Background}\label{sec:related-works}

Considering the limitation of the current text-to-3D generation and the availability of existing datasets of 3D objects, this research instead focuses on improving 3DOR. Quick access to specific 3D objects can significantly enhance creativity and productivity for 3D designers. 3DOR refers to the computational methods to identify and fetch 3D objects within a dataset based on similarity to a query~\citep{He2018TripletCenterLF}. This query, in principle, can take various forms, ranging from complex data structure (multi-view image~\citep{Su2015MultiviewCN}, point cloud~\citep{Zhou2017VoxelNetEL} and voxel~\citep{Wu20143DSA}) to attributes of the items or simple text descriptions. 

In the simplest case, a general-purpose search engine such as \textsc{Google} may be used to retrieve 3D objects from the internet. However, a general-purpose engine relies solely on the labels of the source of the 3D objects and some limited visual cues for retrieval. Also, such an engine has limited access to 3D datasets. In contrast, machine learning (ML)-driven algorithms can compare geometric shapes, textures, and other relevant features of 3D objects to find matches or closely related items~\citep{10.1145/3377876, Feng2023HypergraphBasedMR}. Existing methods are usually designed around computer vision (CV) techniques and thus focus more on the visual characteristics of 3D objects, missing the critical step of understanding the semantic meaning of user input and associating them with visual cues. Various methods have been proposed to enable 3DOR based on either single modality~\citep{Bustos2007ContentBased3O, Wei2020ViewGCNVG, Su2020JointHF} or multiple modalities~\citep{Nie2019MMJNMJ, Feng2023HypergraphBasedMR}, aiming to learn the distribution of 3D objects in high dimensional space~\citep{He2018TripletCenterLF} and sorting the items according to various distance metrics. In a typical setting of closed-set 3DOR, the training set $\mathcal{D}_{train}:=\{(x_i, y_i)\}^N_{i=1}$ and the testing set $\mathcal{D}_{test}:=\{(x_j, y_j)\}^M_{j=1}$ are assumed to have the same distribution and can be represented in the same space, where $x_k$ is a 3D object and $y_k\in \mathcal{Y}:=\{c_l\}^C_{l=1}$ is the associated category of the 3D object. The retrieval method is trained on the training set to optimize: 
\begin{align}
    f^* &= \operatorname{argmin}_{f\in \mathcal{H}} \mathbb{E}_{(\mathcal{D}_{train}, \mathcal{D}_{test})} \nonumber\\
    &\mathbb{I}(y_i=y_j) \big(1 - e^{-\operatorname{dist}(f(x_i), f(x_j))}\big)
    + \mathbb{I}(y_i\ne y_j) \big(e^{-\operatorname{dist}(f(x_i), f(x_j))}\big).
\end{align}
This objective encourages 3D objects in the same category to be closer and objects in different categories farther apart in the representation space. It is also very dependent on the categories. The method learned may be meaningless if the category is very broad. Current 3D datasets are more focused on the representation or modality of 3D objects to capture visual and spatial information, undermining the importance of the semantic meaning of these 3D objects. Some works, such as ShapeNet~\citep{shapenet2015} and Objectverse~\citep{deitke2023objaversexl}, tried to fill in this blank by adding general labeling to 3D objects with limited detail. As a result, 3DOR methods tend to rely on complex visual or 3D input for retrieval. However, in the use case for 3D designers, providing visual representation to retrieve 3D objects is challenging in the ideation stage. A textual description of the 3D object might be the most practical cue to use in the ideation phase of design. In order to enable retrieval based on text with sufficient accuracy, the text label of 3D objects needs to be improved for existing 3D datasets. 

To address this limitation, this research leverages the recent breakthrough in multi-modal large language models (LLMs) to propose an ML-enhanced framework \clas{}, which labels datasets of 3D objects with human-readable descriptions focusing on different perspectives (such as geometry, functionality, or design concept) based on prompts and associates these descriptions with their corresponding visual characteristics in high dimensional vector space for easy retrieval using words. Systems built with \clas{} offer more precision and flexibility in 3D object retrieval by understanding complex spatial and/or design-related language, where traditional methods are not specialized.

\section{Method}\label{sec:framework}

To automate the process of labeling, indexing, and retrieving 3D objects from a dataset, we propose \clas{} - an ML-enhanced four-step (capture, label, associate, and search) framework as illustrated in Figure~\ref{fig:framework}. To demonstrate the framework's effectiveness, we deploy it to create a proof of concept search system for chairs in the \textsc{ShapeNet} dataset~\citep{shapenet2015} as demonstrated in Section~\ref{sec:application}. Applying \clas{} on an existing 3D dataset creates two products: a set of descriptive labeling for the datasets, which can be used for other training or evaluation, and a 3DOR retrieval method based on text input.

\paragraph{Capture.} The first step in \clas{} is to capture a certain number of images of the rendered 3D objects. Since 3D objects can be stored in various formats, the environment or software that could be used to render and capture these objects varies. The cross-platform game engine \textsc{Unity} or 3D computer graphics software \textsc{Blender} are able to cover most of the formats. The setting of the lens, orientations, and lighting should be tuned to suit the purpose of the dataset. The images taken are forwarded to the next step for further processing. This step generates a 2D representation of the 3D objects. Depending on the complexity of the 3D model, more than one image may be generated to preserve more details. In the proof of concept application, we used the images of 3D objects of the chair category in the \textsc{ShapeNet} dataset~\citep{shapenet2015}.  

\paragraph{Label.} In the second step, \clas{} generates a description of the image and labels the image with the corresponding description. Many 3D datasets, such as \textsc{ShapeNet}~\citep{shapenet2015} and \textsc{ModelNet}~\citep{wu20153d}, are released with limited labels, making reusing or repurposing these datasets outside of the research community difficult. This step is performed using a multi-modal LLM, such as \textsc{ChatGPT-4V}, to describe the image with a specific focus defined by the user. For example, the description can be directed to focus on the object's appearance and/or usage. The images of 3D objects and their accompanying descriptions are used in the training in the next step. The performance of the retrieval method is very dependent on the quality of the labeled description. The prompt to generate the descriptions should be designed considering the users of the retrieval system, i.e., what kind of language is usually used to describe the objects. In the proof of concept application, we used \textsc{ChatGPT-4V} with engineered prompts to generate detailed descriptions of the shape and design features of the chairs.  

\paragraph{Associate.} This step trains a multi-modal ML model, i.e., a model capable of encoding two different modalities (text and image), to associate images and their descriptions. Different models can be used for this purpose based on the design of the system, such as \textsc{CLIP}~\citep{radford2021learning} with vision transformer~\citep{dosovitskiy2021image}. After training, two vector spaces representing the textual descriptions and visual images are created, respectively. The similarity of items can be reflected by their proximity in the vector space. We fine-tuned a \textsc{CLIP} model for proof-of-concept application to associate the two modalities of 3D objects.

\paragraph{Search} The final step prepares a user interface (UI) for the users to engage with the system, i.e., search and retrieve 3D models given a textual description. Based on the purpose of the system, different UIs can be used, ranging from a command line to a webpage to an immersive AR/VR environment. We created a website using React to host the proof of concept application. UI ultimately determines how users interact with the AI system. 
\section{Proof of concept application}\label{sec:application}
\begin{figure}[t]
    \centering
    \includegraphics[width=0.99\linewidth]{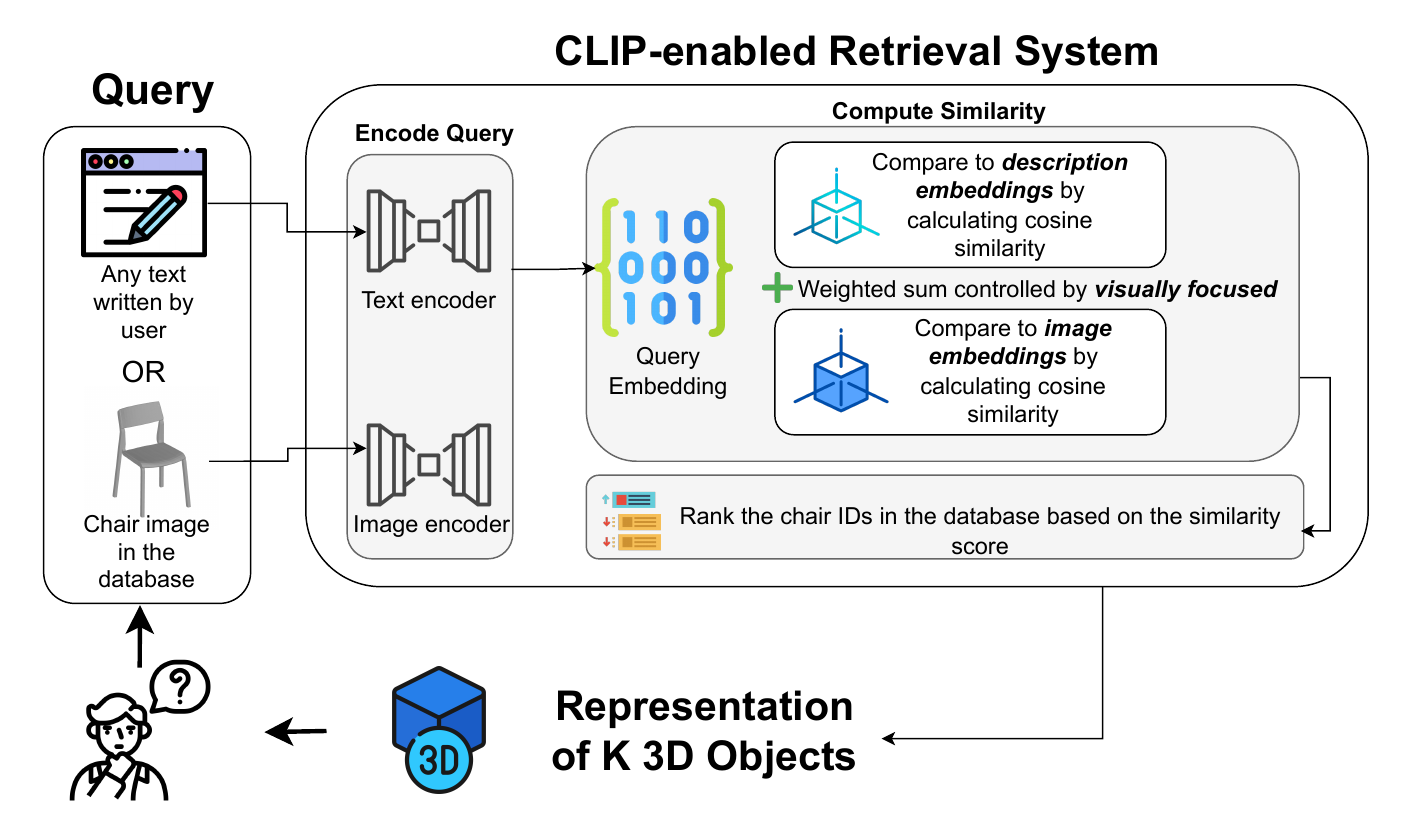}
    \caption{Overview of the \clas{} powered application for retrieving 3D chair objects using natural languages.}
    \label{fig:application}
     \vspace{-12pt} 
\end{figure}

In this section, we present a \clas{} powered web application that allows users to search for 3D objects of chairs using words to demonstrate the usefulness of \clas{}. We first introduce the background of the text-based search application, followed by a detailed explanation of how each step in \clas{} is applied. An overview of this use case is provided in Figure~\ref{fig:application}. Implementation details can be found in Appendix~\ref{appx:web}. The interaction between users and the system is described in details in Appendix~\ref{appx:design}.

\subsection{Background}
Prototyping new designs can be challenging for 3D designers who need references for inspiration, but effectively matching the needs of these designers and existing 3D objects can be challenging. We chose a specific category, i.e., chairs, to demonstrate how \clas{} can be applied to bridge the gap: chair designs. \shapenet{} dataset~\citep{shapenet2015} contains 6,778 3D chair objects, which can be potentially used for creative works. However, it is impractical for designers to manually go through these objects, considering that they often lack proper annotations that are easily readable by humans. An exemplary use case by designers is shown in Appendix~\ref{appx:design}.

\subsection{Implementing 3D chair search system using \clas{}}
This system was built using \clas{} as detailed below. Note that we leveraged the existing images in the dataset so the first step in \clas{}, i.e., capture, is skipped.

\subsubsection{Label} 

\begin{figure}[t]
    \centering
    \includegraphics[width=\linewidth]{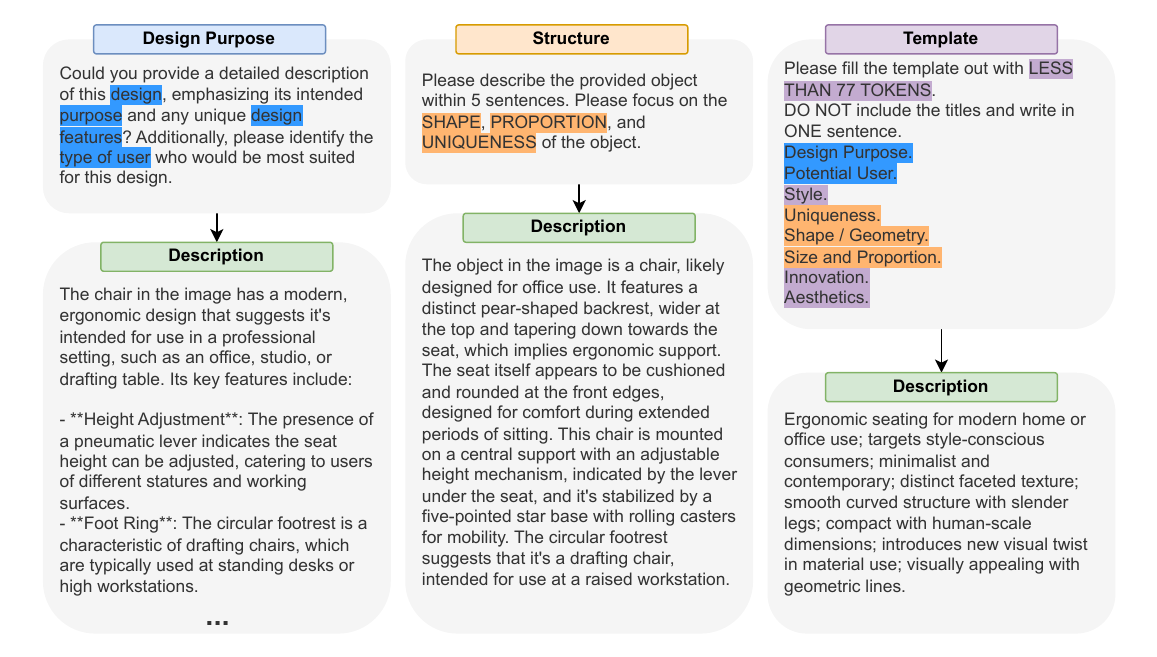}
    \caption{Example of different prompt designs. The generated description varies according to the prompt given the same chair image.}
    \label{fig:prompts}
\end{figure}

We conducted prompt engineering for the \textsc{ChatGPT-4V} to generate image descriptions. The generated descriptions are prompted to match potential user input that designers would likely use when searching for chair objects. We design prompts echoing three focuses to achieve the goal: design purpose, structure, and template.

\paragraph{Design purpose.}
Product design often starts from the product's functionality to solve a specific problem and innovate from the original design. From the designers' perspective, potential users may enter the design purpose and use case as input prompts to get ideas for the brainstorming process. The left column in Figure~\ref{fig:prompts} shows a prompt containing the key questions for designers to generate a design purpose-based text description.

\begin{figure}[t]
  \centering
  \includegraphics[width=0.85\linewidth]{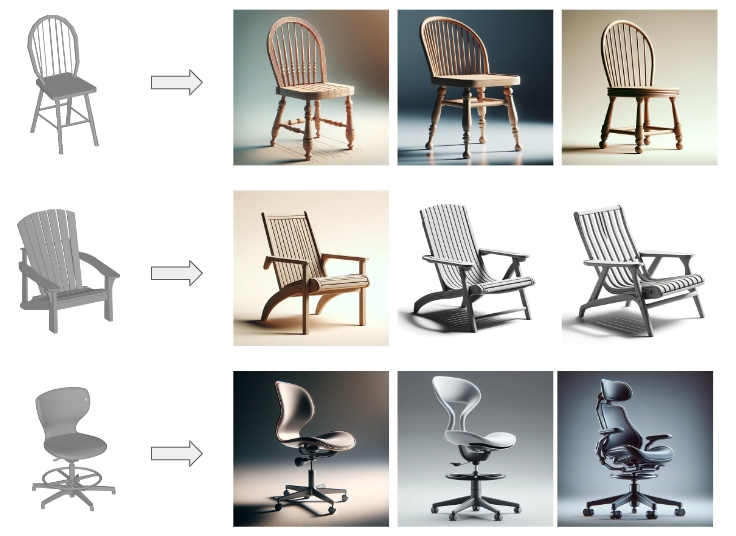}
  \caption{Effectiveness of prompt \textit{Structure}. The images on the left are fed to ChatGPT-4V with the prompt, and the generated text descriptions are then fed to DALLE-E to generate the right images.}
  \label{fig:structure-prompt}
\end{figure}

\paragraph{Structure.}
Visual characteristics play an important role in distinguishing different chairs and are critical in making design decisions. A detailed description of the exact shape, uniqueness, and size relations would be very helpful for the retrieval method later to learn different 3D objects with enough precision. To discover the effective prompt, we started with the simple prompt, ``\textit{Can you describe the chair’s appearance?}"  The optimization process of the prompt underwent iterative tests by adjusting the task description, providing hints, varying the focused aspects, etc. The generated descriptions were initially read and validated manually to omit failed prompts. Then, we evaluated the prompts by comparing the image generated by \textsc{DALL-E} from OpenAI to the original image. The more similar the generated image using the description to the original image, the better the prompt is. Several examples of tested prompts and corresponding results are included in Appendix~\ref{appx:prompt}. Based on the experiments, capitalizing the critical words and adding a sentence to highlight the unnecessary part for the description would increase the performance of the prompt. On the other hand, prompts with too much information and the use of general words might produce imprecise descriptions of the provided objects. The conciseness of the generated description is important to avoid overly general expressions and highlight the characteristics of an object. The finalized structure prompt is shown as the middle column in Figure~\ref{fig:prompts}.

\paragraph{Template:}
Lengthy descriptions are expected to contain more information about the chair. However, they also mean more computation is needed to train and serve the retrieval system. In this proof of concept application, the base version of the \textsc{CLIP}~\citep{radford2021learning} was used, limiting the length of the generated description to less than 77 tokens. To compress the description's size while maintaining each chair's distinctive features, we used a template-like prompt as the right column in Figure~\ref{fig:prompts}. It focuses on both design purpose and structure and is used for the labeling process. The template can also be provided to the user for filling out during searching, which could help to establish consistency between the generated descriptions and potential user input to improve the accuracy of the retrieval system. Even though the limited structural descriptions generated by the template did not cover all the details of the complex 3D objects, they maintained the advantage of describing the design purpose and uniqueness.  

\subsubsection{Associate} 

\begin{figure}[t]
    \centering
    \includegraphics[width=0.75\linewidth]{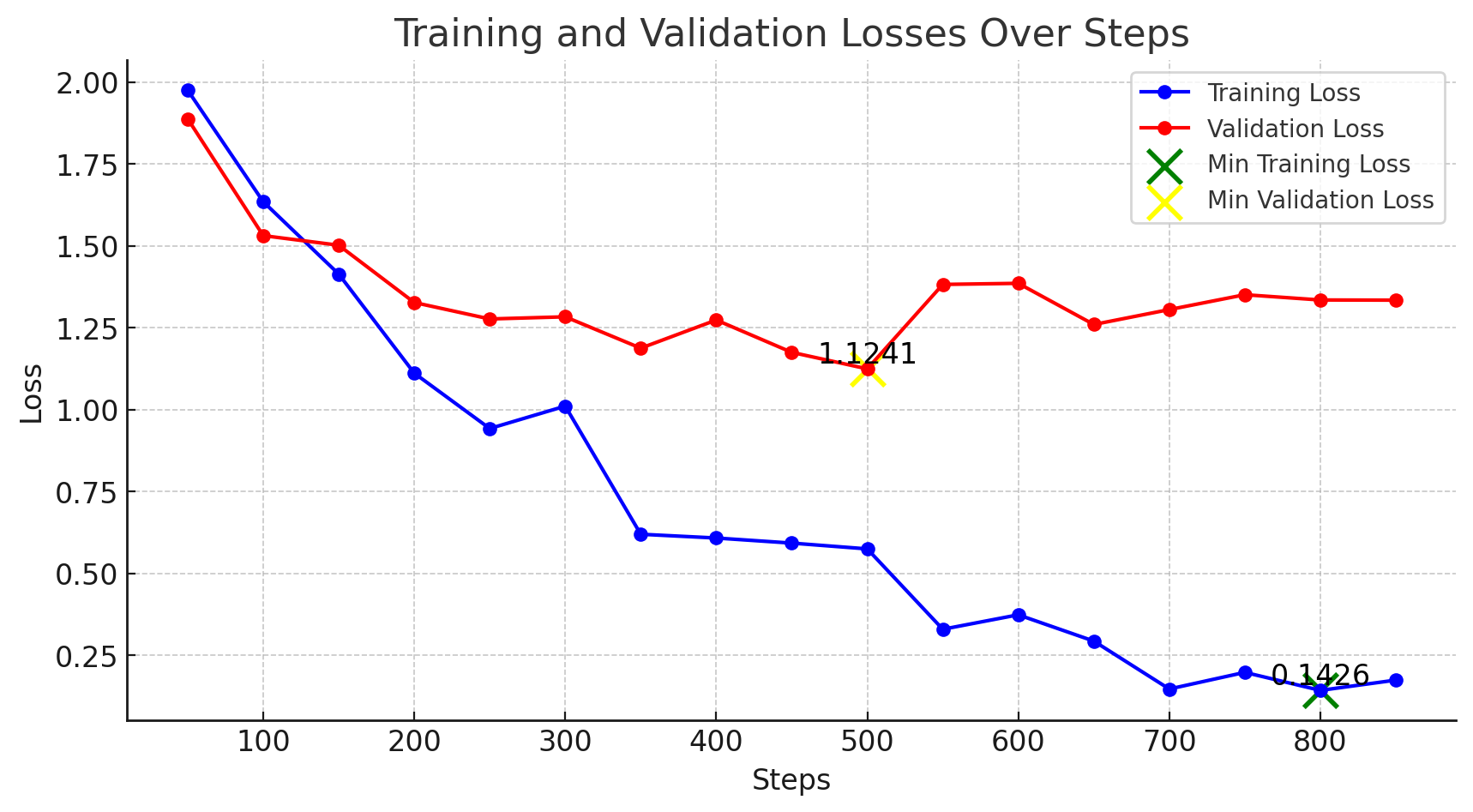}
    \caption{Fine-tuning of the CLIP model. Batch size: 32. Epochs: 5. Weight decay: 0.01. Warm-up steps: 50. Learning rate (with cosine scheduler): 2e-5. The pairs of images and descriptions in the validation set are not seen by the model in training, which is unlikely to happen in close-set retrieval as the model is only used to retrieve items in a fixed dataset. The fine-tuning took about 30 minutes on 80\% (in which 80\% for training and 20\% for validation) of the chairs in the ShapeNet dataset using an NVIDIA Tesla P100 GPU.}
    \label{fig:fine-tuning}
    \vspace{-20pt}
\end{figure}

We used \textsc{CLIP}~\citep{radford2021learning} to associate the text descriptions and the image used to generate the text description. The architecture of the \textsc{CLIP} is included as Appendix~\ref{appx:clip}. We first produced the datasets for fine-tuning and evaluation by pairing up images of 3D chair objects and their corresponding text descriptions generated by \textsc{ChatGPT-4V} prompted by the \textit{template} design. Pairs in the training set were fed into the CLIP model, and the fine-tuning encouraged the text embedding embedded by the text encoder and the image embedding embedded by the image encoder to be close together after projection into the vector space of the same dimension by the contrastive loss:

\begin{align}
\mathcal{L} = - \frac{1}{N} \sum_{i=1}^{N} \left( \log \frac{\exp \left( \text{sim}(x_i, y_i) \right)}{\sum_{j=1}^{N} \exp \left( \text{sim}(x_i, y_j) \right)} + \log \frac{\exp \left( \text{sim}(y_i, x_i) \right)}{\sum_{j=1}^{N} \exp \left( \text{sim}(y_i, x_j) \right)} \right)
\end{align}

where $N$ is the number of samples in the batch, and $\text{sim}(x, y)$ is the cosine similarity between the image representation $x$ and the text representation $y$. The labels are implicitly indicated by the pairs $(x_i, y_i)$ which are the correct pairs in the batch, while $(x_i, y_j)$ and $(y_i, x_j)$ for $j \neq i$ represent the negative pairs. The fine-tuning is visualized as Figure~\ref{fig:fine-tuning}. We noticed that the model might overfit and did not generalize well, which is discussed in Section~\ref{subsec:eval}. After the fine-tuning, the CLIP model became more sensitive to chairs and their designs than a general model and could be used for retrieval.

\subsubsection{Search}

\begin{figure}[ht]
    \centering
    \begin{subfigure}[b]{0.49\linewidth}
        \centering
        \includegraphics[width=\linewidth]{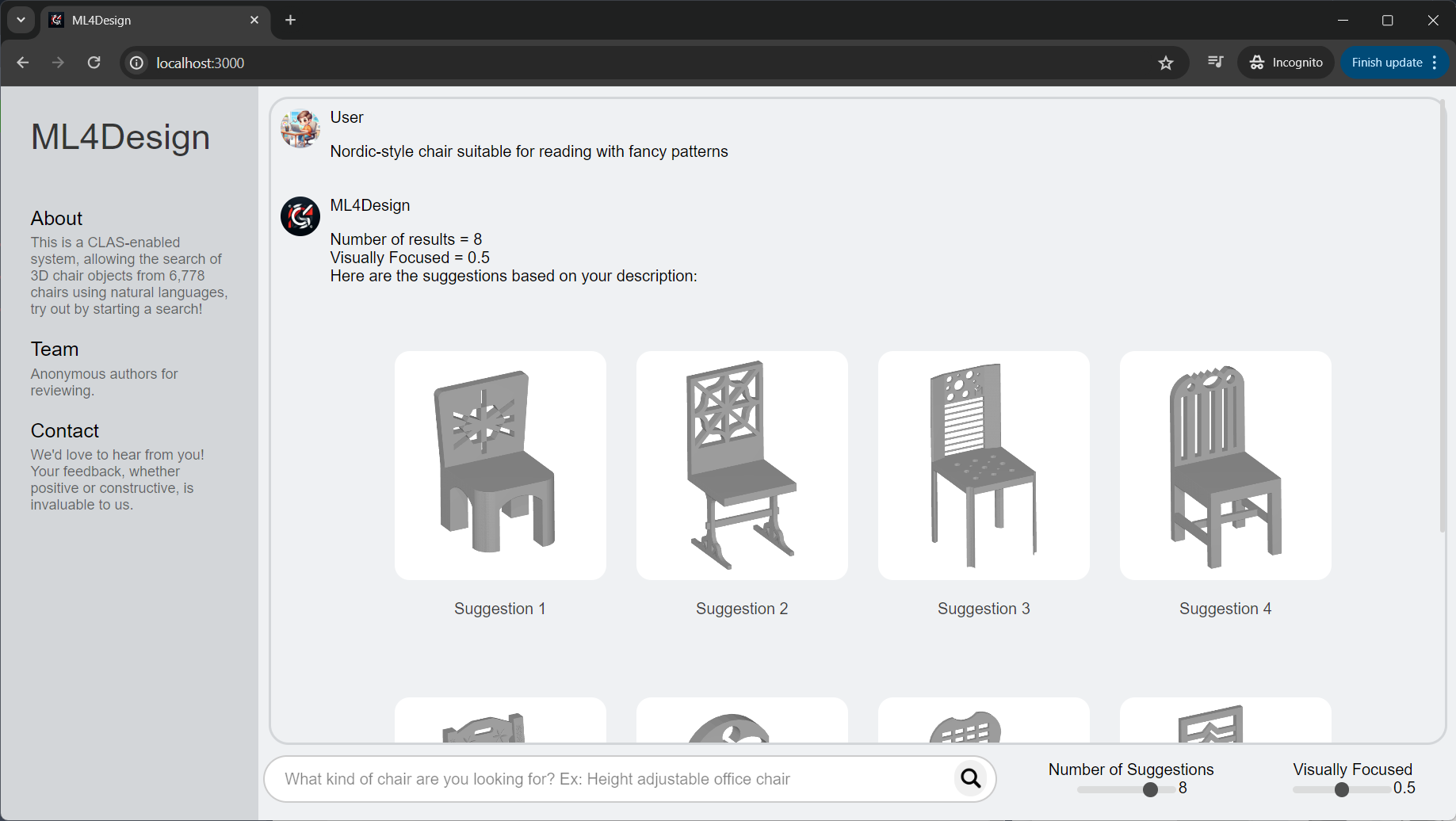}
    \end{subfigure}
    \hfill
    \begin{subfigure}[b]{0.49\linewidth}
        \centering
        \includegraphics[width=\linewidth]{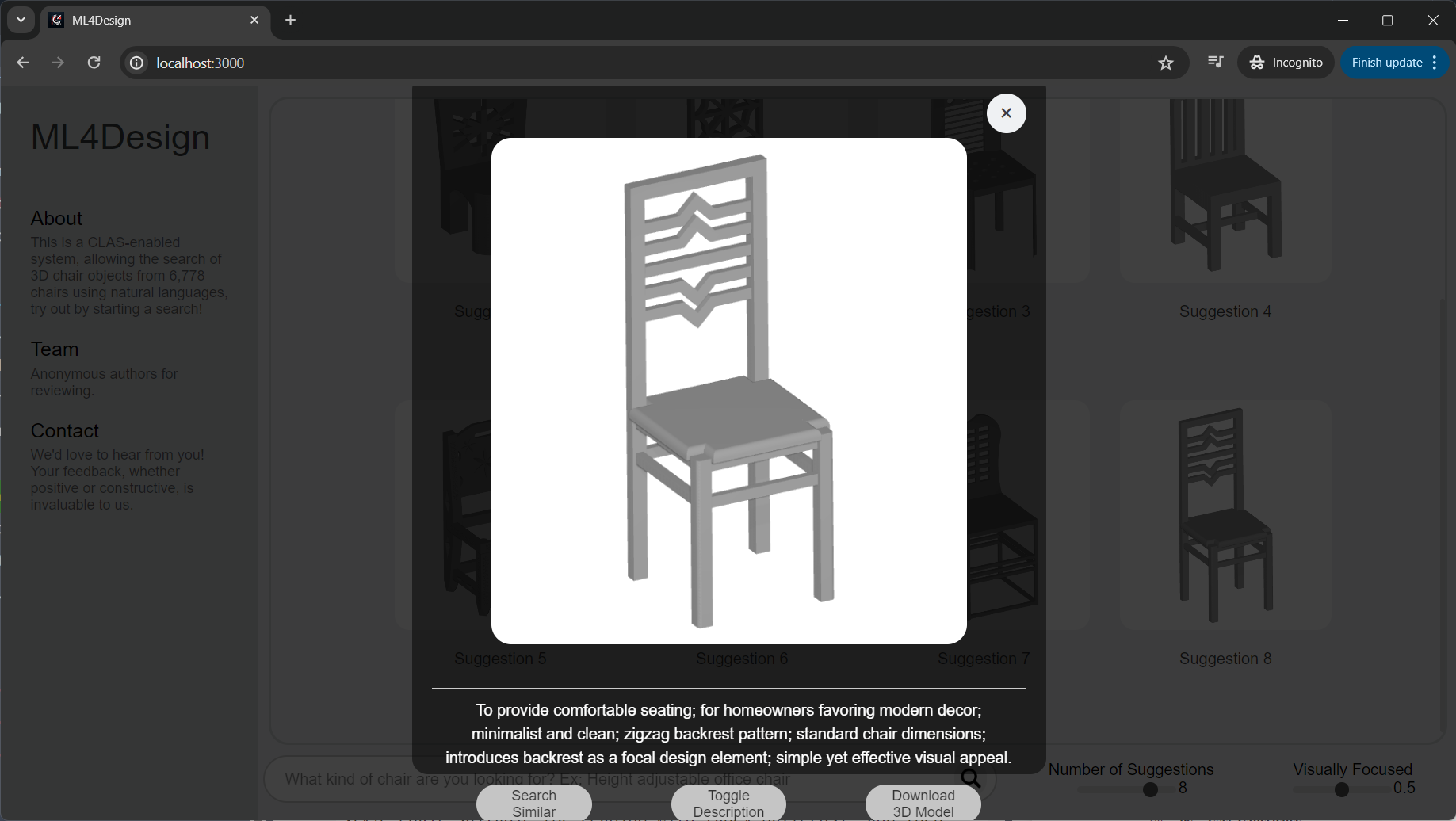}
    \end{subfigure}
    \caption{\textit{Left}: example of searching chairs using natural languages. The query entered is ``\textit{Nordic-style chair suitable for reading with fancy patterns}" and the ``\textit{Number of results}" is set to 8. \textit{Right}: Enlarged picture after selecting a chair.}
    \label{fig:search-example}
\end{figure}

A complete web application with a front end and a back end was created to host the search system. The implementation details are included in Appendix~\ref{appx:web}. When designers start a new search, the process begins with the search bar, where they enter a prompt such as ``\textit{Nordic-style chair suitable for reading with fancy patterns}" and then select the value for two slide bars, pressing either enter or the search button to search. Upon submitting the prompt, our system quickly retrieves 3D objects, displaying search results within a fraction of a second from a database of 6,778 chair objects. The search results in the chat window are presented in rows, with the user's avatar displayed above, showing the user's prompt, followed by the system's response, ``\textit{Here are the suggestions}," to enhance user interaction (as demonstrated in Figure~\ref{fig:search-example}). As users browse through the objects, the image of the object they hover over will slightly enlarge, indicating which chair object they are considering or wish to select. Clicking on a desired model enlarges it for a closer inspection of the details (as shown in Figure~\ref{fig:search-example}).  To download the chosen 3D object, clicking the download button redirects the user to object downloads. If the user is unsatisfied with the search results, they can click on the ``\textit{show description}" button to view the label/description generated by \textsc{ChatGPT-4V} for each model. This feature allows users to refine their search based on the labels or start a new search by entering a more precise prompt. If the user wants to search for more 3D objects that are similar to one of the result models, simply click on ``\textit{search similar}"; the website will search again using the selected object as a query, thereby providing similar object suggestions. These functions offer a tailored and interactive browsing experience.

This website was designed for designers seeking inspiration and resources for 3D object designs. The visual and workflow designs help users focus more on the search system. The pre-filled prompt in the search bar enables users to quickly start their initial search, while the navigation bar offers additional information and content about the project. Most importantly, the search results page makes the search process more intuitive and convenient, allowing users to easily find and download the object they seek, leaving a good impression and a great user experience.

\begin{figure}[t]
    \centering
    \includegraphics[width=1\linewidth]{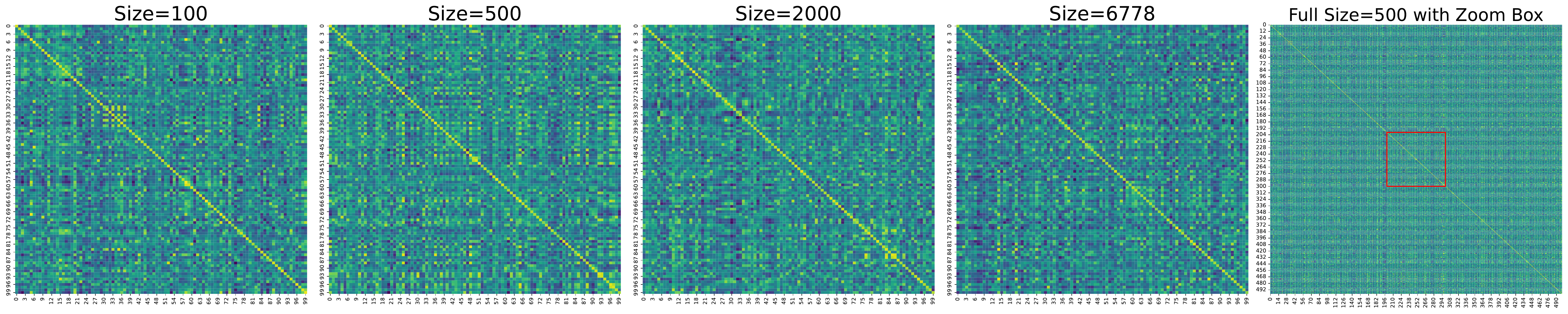}
    \caption{Similarity matrix of text embeddings and image embeddings as a heat map. The size of the dataset to perform retrieval is 100, 500, 2000, and 6778 (full dataset), accordingly. For better representation, the 100 chair objects in the middle are selected to present as the red box in the rightmost heat map. A clear diagonal line can be seen in all cases, indicating a high similarity between the text and image representations of a chair object.}
    \label{fig:heatmap}
\end{figure}

\subsection{Evaluation}\label{subsec:eval}
\begin{table}[t]
    \setlength{\tabcolsep}{6pt}
    \scriptsize
    \centering
    \caption{
        Experimental results achieved by fine-tuning the training set or on the complete set (close-set retrieval set-up). Results were achieved using the CLIP model with the best train loss and validation loss, respectively. Baseline refers to the CLIP model by OpenAI without fine-tuning. 
    }
    \begin{tabular}{ccccccccccc}
        \toprule
        & \multirow{2}{*}{Size} & \multicolumn{3}{c}{MRR (0-1)} & \multicolumn{3}{c}{Top-1 Acc (\%)} & \multicolumn{3}{c}{Top-10 Acc (\%)} \\
        \cmidrule(lr){3-5} \cmidrule(lr){6-8} \cmidrule(lr){9-11}
        & & Baseline & Best Tr. & Best Val. & Baseline & Best Tr. & Best Val. & Baseline & Best Tr. & Best Val. \\
        \midrule
        Train & 5422 & 0.02 & \textbf{0.57} & 0.20 & 0.57 & \textbf{41.53} & 9.94 & 3.50 & \textbf{88.68} & 42.05 \\
        Test & 1365 & 0.05 & \textbf{0.24} & 0.18 & 2.36 & \textbf{11.95} & 9.37 & 7.96 & \textbf{47.64} & 38.20 \\
        Complete & 6778 & 0.02 & \textbf{0.45} & 0.16 & 0.55 & \textbf{31.53} & 7.73 & 3.24 & \textbf{72.68} & 33.60 \\
        \bottomrule
        \\
        \textit{Close-set} & 6778  & 0.02 &\multicolumn{2}{c}{\textbf{0.58}} & 0.55 &\multicolumn{2}{c}{\textbf{42.27}} & 3.24 &\multicolumn{2}{c}{\textbf{89.64}}\\
        \bottomrule
    \end{tabular}
    \label{tab:results}
        \vspace{-10pt}

\end{table}

Before the \clas{} implementation, no efficient solution was available to retrieve chairs in the \shapenet{} using natural languages. Potential users must go through all the chairs to manually decide which are wanted, which requires much time and effort. To the best of our knowledge, this exemplary use case is the first attempt to enable this convenient search using natural language on 3D chair datasets. 

The out-of-box CLIP model~\citep{radford2021learning} from OpenAI is used as the baseline to demonstrate the effectiveness of the fine-tuning using three standard metrics in evaluating retrieval systems. \emph{Mean reciprocal rank (MRR)} calculates the mean of the reciprocals of the rank at which the true objects are retrieved. MRR takes a value between 0 and 1 such that the closer to 1, the better the performance, i.e., the system ranks the intended item higher in all the possible items. \emph{Top-1 accuracy} calculates the percentage of retrieved objects as the true object. \emph{Top-$10$ accuracy} expands on the top-10 accuracy to the scope of the top 10 most probable objects selected by the model.

As shown in Table~\ref{tab:results}, there is a huge increase in performance in all the metrics for both the training, testing, and complete sets. The fine-tuned model with the best train loss performs noticeably better than the best validation model, even for unseen samples in the test set. We argue that for close-set retrieval, where objects to be retrieved are from a fixed dataset, it is reasonable to train the retrieval model on all data. We, therefore, overfitted the model on the complete set to achieve a loss of 0.02 in 10 epochs. The overfitted model achieved the best retrieval results, as shown in the last row of Table~\ref{tab:results}. As demonstrated in Figure~\ref{fig:heatmap}, the text and image representation of the 3D chair objects can be closely associated even with 6,778 samples.

\section{Discussion}\label{sec:discussion}
\clas{} targets the problem space around the inefficiencies in finding relevant 3D objects based on user specifications, which often hinder the design process. Recognizing the importance of facilitating designers in finding inspiration and overcoming creative blocks, we aim to enable efficient usage of existing 3D datasets. We propose an ML-enhanced framework \clas{} to enable effective 3D datasets indexing, labeling, and prompt-driven retrieval. Compared to existing methods, \clas{} differentiates itself by offering incredible flexibility in the 3D datasets, which operate on and prompts with different levels of details and focus, filling a gap left by generative models and general search engines. 

\subsection{Limitations}\label{subsec:limitation}
\paragraph{Dealing with potentially much larger datasets with hybrid categories.}
In the proof of concept application, we tested \clas{} on a dataset of 6,778 chairs. However, various methods can be experimented with to expand the capacity of the current formulation of \clas{}. For instance, one may add a classifier model to classify the 3D objects into different categories and forward them to corresponding CLIP models (assuming there are multiple CLIP models for multiple categories). One may also increase the length of descriptions and the dimension of embedding to encourage the CLIP model to learn the complex relation directly, assuming there is enough data. 

\paragraph{Enhancing understanding of negative prompts.}
The current search function provided by \clas{} struggles to understand negative prompts and often does the opposite. For example, if a user states not to retrieve objects with certain properties, the system would actively pick objects with unwanted properties. This occurs primarily as no negative example exists in training for retrieval systems. The text description generated for each image usually highlights what the object has rather than does not have. To fix this, one may change the prompt to generate different descriptions or change searching in the embedding space to substrate the embedding of features that are not wanted. More in-depth study is required as negative prompts are usually not supported by retrieval.
 
\paragraph{Enhancing support for retrieval of sets of items or scenes.}
Currently, the \clas{}-powered system only retrieves a single subject in each suggestion. However, it would be beneficial to many designers if the application could understand the relation between objects of different categories and combine them while evaluating the overall design effect. Future developments could focus on integrating different items to form sets or scenes, allowing designers to gain more inspiration and obtain a comprehensive overview of their designs. This function would enable users to explore various combinations of different items and evaluate designs more effectively.

\subsection{Social impact}\label{subsec:impact}
\clas{} enables any person or organization to repurpose and/or reuse their existing 3D datasets effectively. \clas{}-processed 3D dataset allows users to search its content using natural languages with high accuracy and precision. We hope that exposing numerous references to designers when designing new concepts can assist them in ideation. We demonstrate the usefulness of \clas{} in powering the search for 6,778 3D chairs, which can be used by furniture designers. Similarly, \clas{} can be applied to different 3D datasets for different purposes. By adjusting the prompt used to generate text descriptions for images of 3D objects, the search function can be tuned to work with different aspects of the objects suiting different use cases. By bridging the gap between user prompts and relevant 3D models, we aspire to enhance the design process and foster creativity within the community.

\begin{ack}
    We sincerely thank the support of our project advisor, Professor Kosa Goucher-Lambert, offered throughout the project. We would also like to thank Kevin Ma in the lab and Daniele Grandi from Autodesk for regular meetings and feedback on our project. It would not be possible for us to complete this project as it is without the help of anyone mentioned above.
\end{ack}

\small
\bibliographystyle{plainnat}
\bibliography{custom}
\newpage
\appendix
\section{Web application}\label{appx:web}

\begin{figure}[h]
    \centering
    \includegraphics[width=\linewidth]{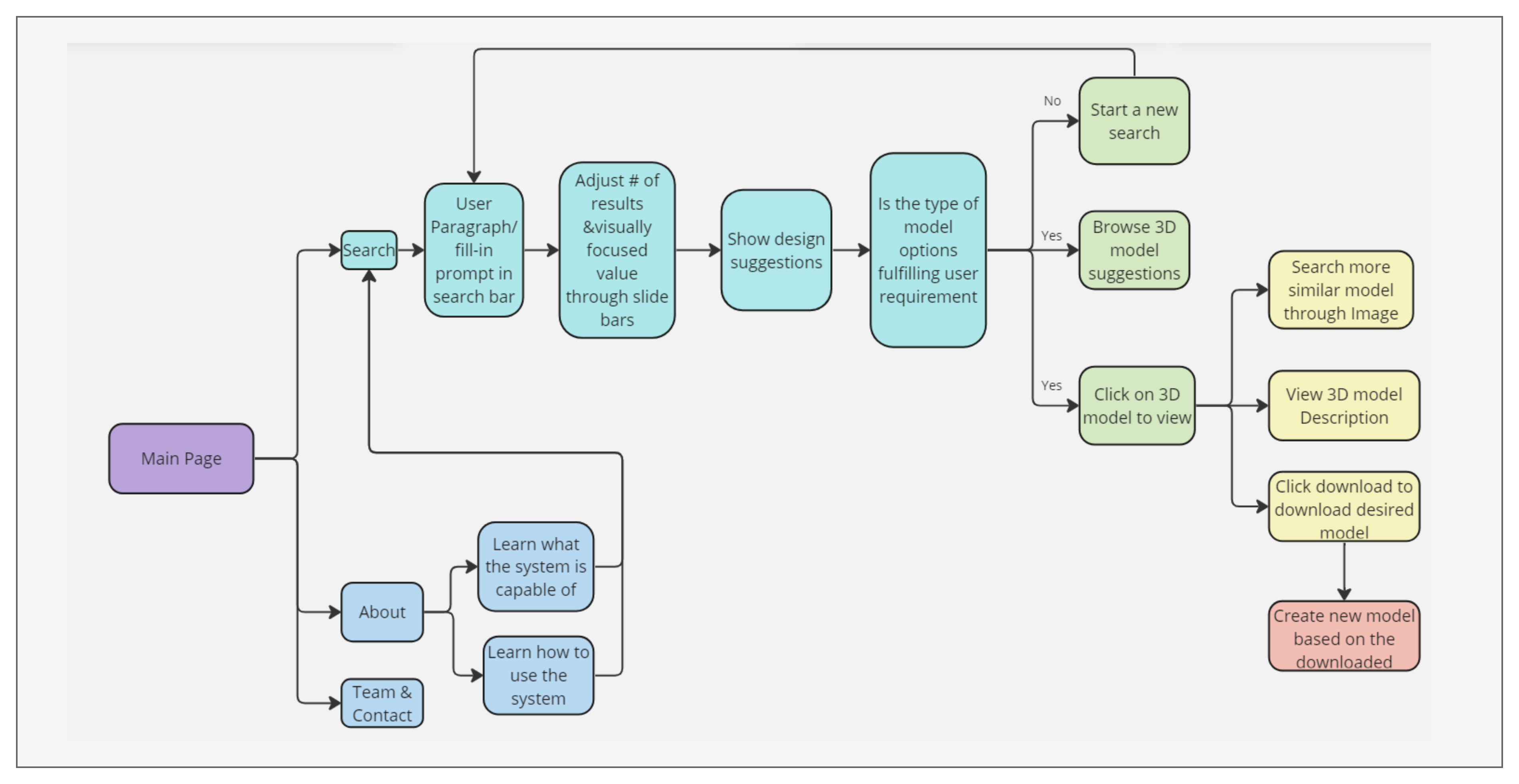}
    \caption{Interaction between the user and the application.}
\end{figure}

\begin{figure}[h]
    \centering
    \includegraphics[width=\linewidth]{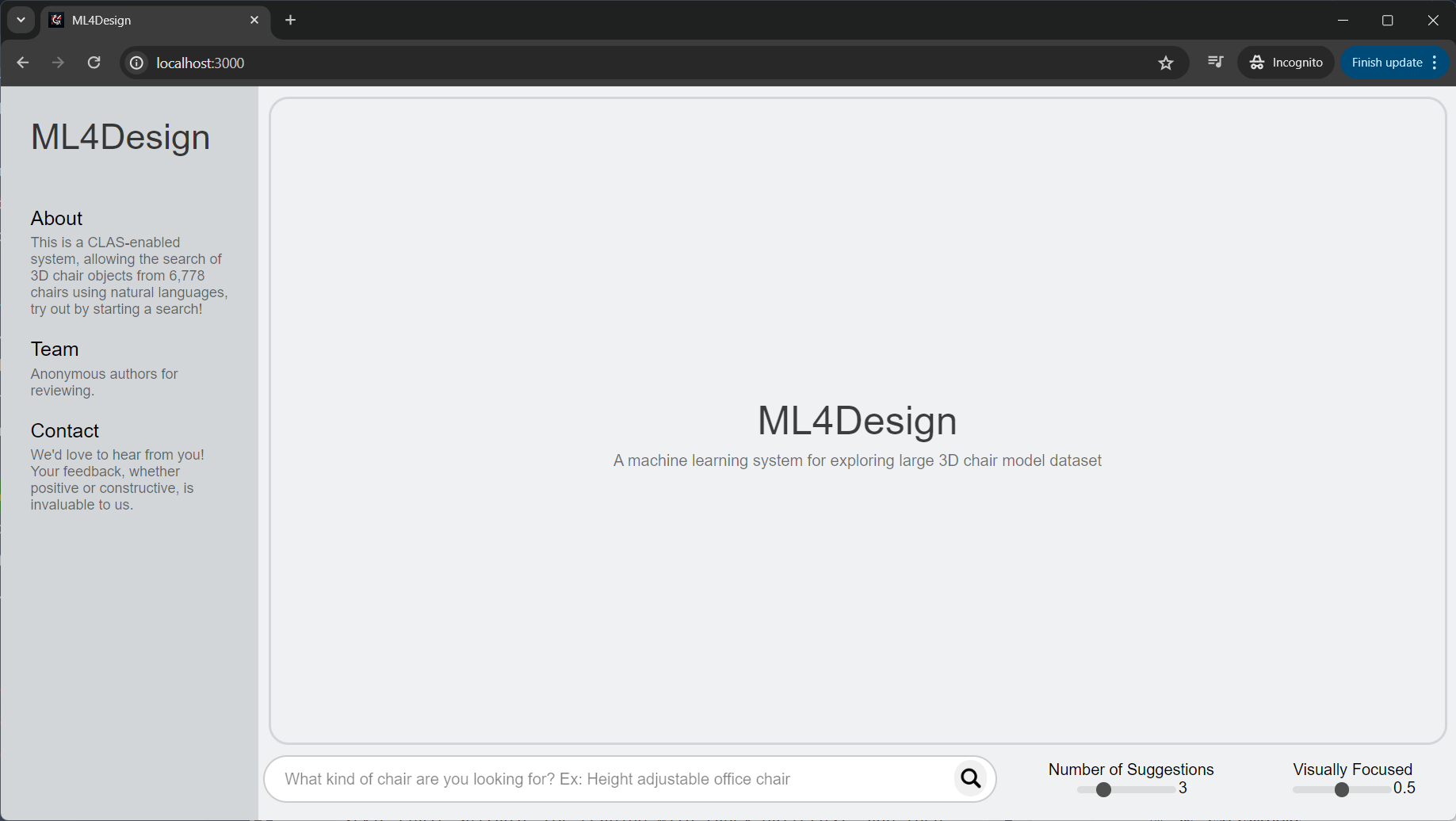}
    \caption{Homepage of the web application.}
    \label{fig:enter-label}
\end{figure}

This web application will be made available to collect feedback.
\subsection{Implementation details}
\paragraph{Front end:}
Our website's design aims to offer an intuitive user interface for designers seeking 3D model design suggestions, and prioritizing search functionality to minimize distractions. The main color theme is light gray, with the central space occupied by a chat window—where users interact most with our website (as shown in Figure~\ref{fig:enter-label}). At the center of this chat window is our logo, ``\textit{ML4Design}," with a subtitle, ``\textit{A machine learning system for exploring a large 3D chair model dataset}," which explains our website's purpose. Below the chat window is the search bar, filled with ``\textit{What kind of chair are you looking for?}" in light grey font color to hint the user at our service. An example, ``\textit{Height adjustable office chair}," is immediately provided to guide users on how to engage with our site to encourage interaction. To the right of the search bar is a slider labeled ``\textit{Number of results}." Users can adjust this slider to select the desired number of results, allowing for focused attention on a few suggestions or a broader exploration of up to ten recommended models. The ``\textit{Visually Focused}. Slide bar allows the user to adjust the weighting of image search and text search. Our website can search in the image vector space and the text vector space. For instance, toggling the visual focus value to 0.1 means the image search will weigh 10 percent and the text search will weigh 90 percent of the result. These features enhance the interactive and engaging experience of our website.

\paragraph{Back end:}
 The back end coordinates the CLIP model, which is hosted online using Huggingface's endpoint service, and the database, which is hosted online using AWS service. When a user enters a query into the website, the input is processed and sent to the CLIP model endpoint, which returns the IDs of the retrieved 3D chair objects to the back end. The back end then fetches the corresponding models from the database based on the received IDs and forwards them to the front end for display. 

 \paragraph{Dataset:}
 Chairs in \shapenet{} are stored as objects and images, i.e., every 3D chair object is paired with an image of the chair. We manage this dataset of chairs using Amazon Web Services (AWS) cloud services for optimized storage and accessibility. Image and 3D object files are stored in AWS Simple Storage Service (S3), providing reliable and scalable cloud storage. Metadata, including file names, image links, 3D object download links, and corresponding descriptions, are logged in a MySQL relational database. This database is hosted on AWS Relational Database Service (RDS), ensuring stable and efficient data handling and enhancing data accessibility.

 \subsection{User Experience}

\begin{figure}[!h]
    \centering
    \includegraphics[width=\linewidth]{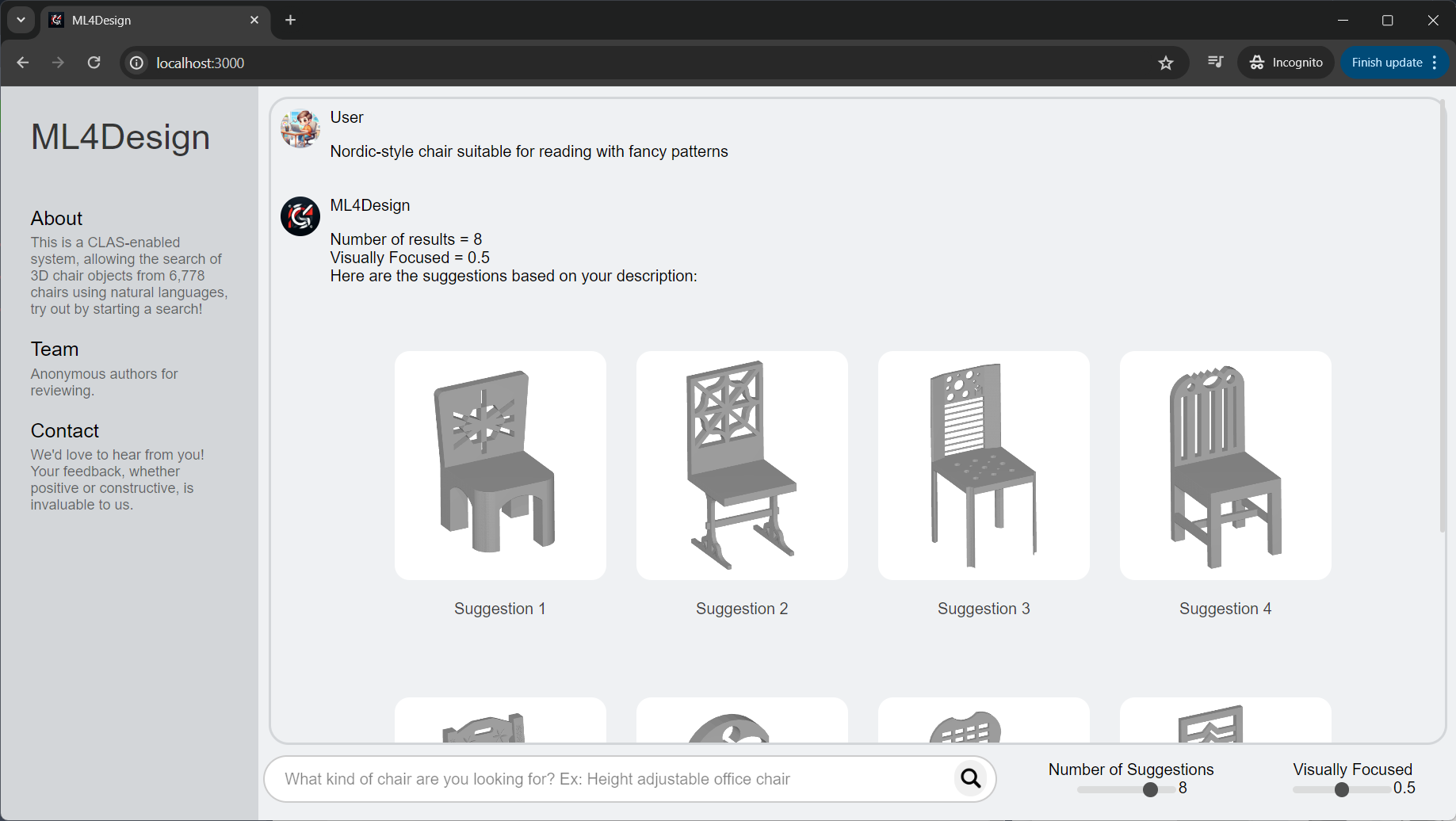}
    \caption{Example of searching chairs using natural languages. The query entered is ``\textit{Nordic-style chair suitable for reading with fancy patterns}" and the ``\textit{Number of results}" is set to 8.}
    \label{fig:search-example-2}
\end{figure}

\begin{figure}[!h]
    \centering
    \includegraphics[width=\linewidth]{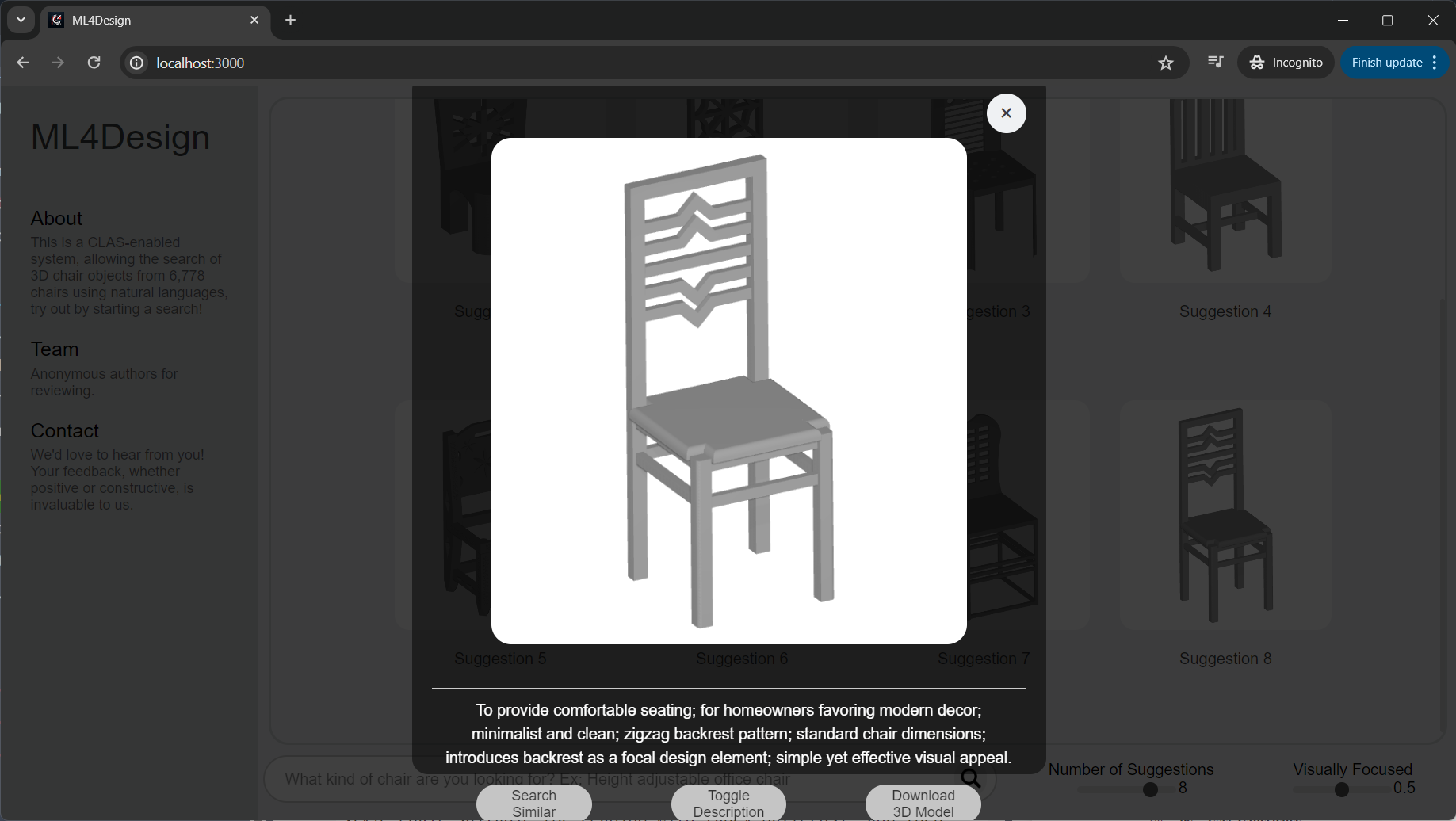}
    \caption{Enlarged picture after selecting a model.}
    \label{fig:enlarge_model-2}
\end{figure}

When designers start a new search, the process begins with the search bar, where they enter a prompt such as ``\textit{Nordic-style chair suitable for reading with fancy patterns}" and then select the value for two slide bars, pressing either enter or the search button to search. Upon submitting the prompt, our system quickly retrieves 3D objects, displaying search results within a fraction of a second from a database of 6,778 chair objects. The search results in the chat window are presented in rows, with the user's avatar displayed above, showing the user's prompt, followed by the system's response, ``\textit{Here are the suggestions}," to enhance user interaction (as demonstrated in Figure~\ref{fig:search-example-2}). As users browse through the objects, the image of the object they hover over will slightly enlarge, indicating which chair object they are considering or wish to select. Clicking on a desired model enlarges it for a closer inspection of the details (as shown in Figure~\ref{fig:enlarge_model-2}).  To download the chosen 3D object, clicking the download button redirects the user to object downloads. If the user is unsatisfied with the search results, they can click on the ``\textit{show description}" button to view the label/description generated by \textsc{ChatGPT-4V} for each model. This feature allows users to refine their search based on the labels or start a new search by entering a more precise prompt. If the user wants to search for more 3D objects that are similar to one of the result models, simply click on ``\textit{search similar}"; the website will search again using the selected object as a query, thereby providing similar object suggestions. These functions offer a tailored and interactive browsing experience.


On the left side of the website, the top of the navigation bar displays our logo, ``ML4Design," followed by three subtitles: ``About," ``Team," and ``Contact." Below these subtitles is a brief description summarizing the content on these pages. The description under the "About" title reads, ``This is a \clas{} framework system, enabling the search of 3D chair models from a dataset of 6,778 chairs. Try it out by starting a search!" This succinctly describes what our system is and its capabilities. Upon clicking on the About page, users can see a slideshow demonstrating how our system's \clas{} framework operates. Below, the slideshow showcases different prompts, offering search examples using various criteria, including searching with chair geometry, design purpose, or a combination of both.

Back to the Navigation bar, the ``Team" page introduces our project members. The ``Contact" page communicates our openness to suggestions and opinions, allowing users to leave feedback. This structure ensures that visitors can easily understand the purpose of our system, learn about the team behind it, and engage with us directly. After viewing these pages, by clicking the top left ML4design logo, the user will be guided back to start their search.

\clearpage
\newpage
\section{Interaction between 3D designers and the system}\label{appx:design}

In this section, we discuss how designers can benefit from a search system enabled by \clas{} and the resources it provides. First, we briefly introduce the design thinking method proposed by Tim Brown ~\citep{brown2008design}. This method later evolved into the 5-step design thinking approach proposed by IDEO, also known as the Double Diamond. This method outlines the process a designer, such as a product designer or 3D model designer, goes through when designing. The first step is ``\textit{empathize}," where designers must understand people's problems and identify issues even if they are not immediately apparent to the users. The second step is ``\textit{define}." In this phase, designers must define the core problem and specifically define what they aim to solve using the POG (Product Opportunity Gap) statement. The third step is ``\textit{ideate}," where a \clas{}-enabled system becomes particularly valuable. In this phase, designers generate and brainstorm solutions for the identified problem. We discuss this in more detail in the following paragraph. The last two steps are ``\textit{prototyping}" and ``\textit{testing}", where designers bring their solutions into the real world and test their effectiveness through user interviews and other methods. Through several iterations, they refine their designs to arrive at the final product.

For a retrieval system powered by \clas{}, the primary goal is to assist 3D model designers during the ``\textit{ideate}" process and the ``\textit{prototyping}" phase. To help understanding, we consider a simple persona and a specific scenario to introduce the entire process the target designers will experience.

\begin{figure}[h]
    \centering
    \includegraphics[width=1\linewidth]{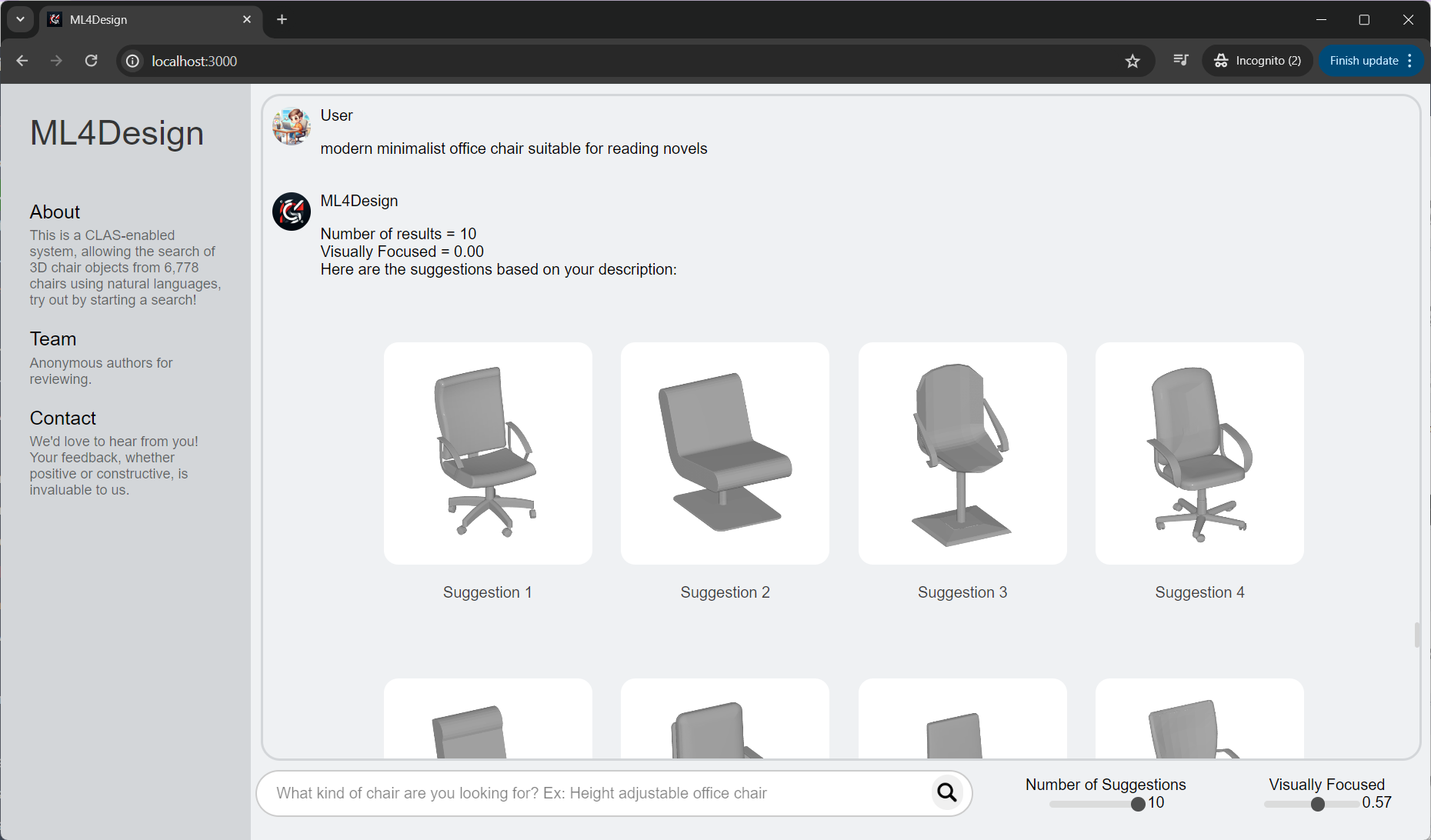}
    \caption{Exemplary use case by Peter. Initial search.}
    \label{fig:story-1}
\end{figure}

\begin{figure}[h]
    \centering
    \includegraphics[width=1\linewidth]{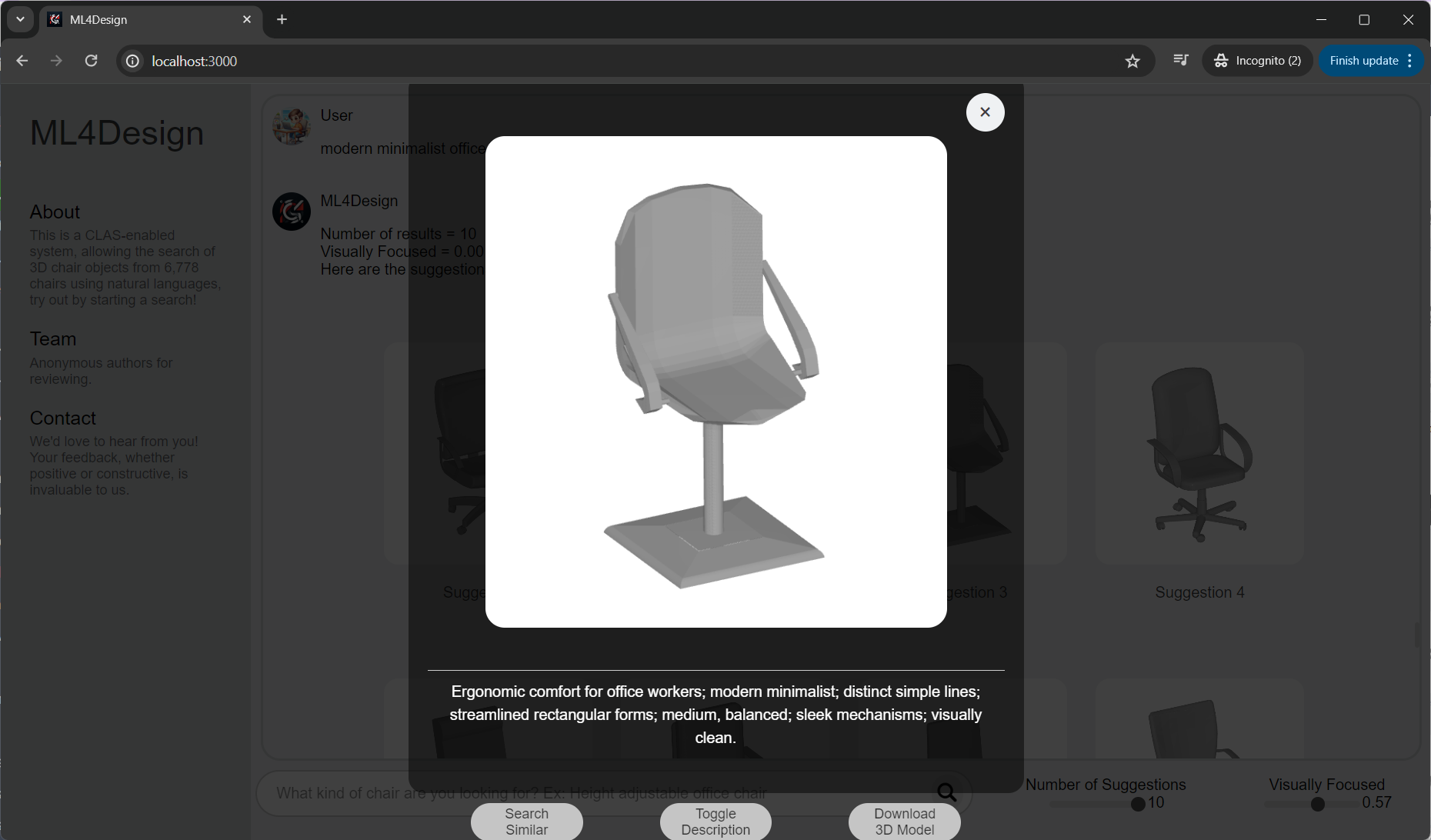}
    \caption{Exemplary use case by Peter. Check one of the suggestions.}
    \label{fig:story-2}
\end{figure}

\begin{figure}[h]
    \centering
    \includegraphics[width=1\linewidth]{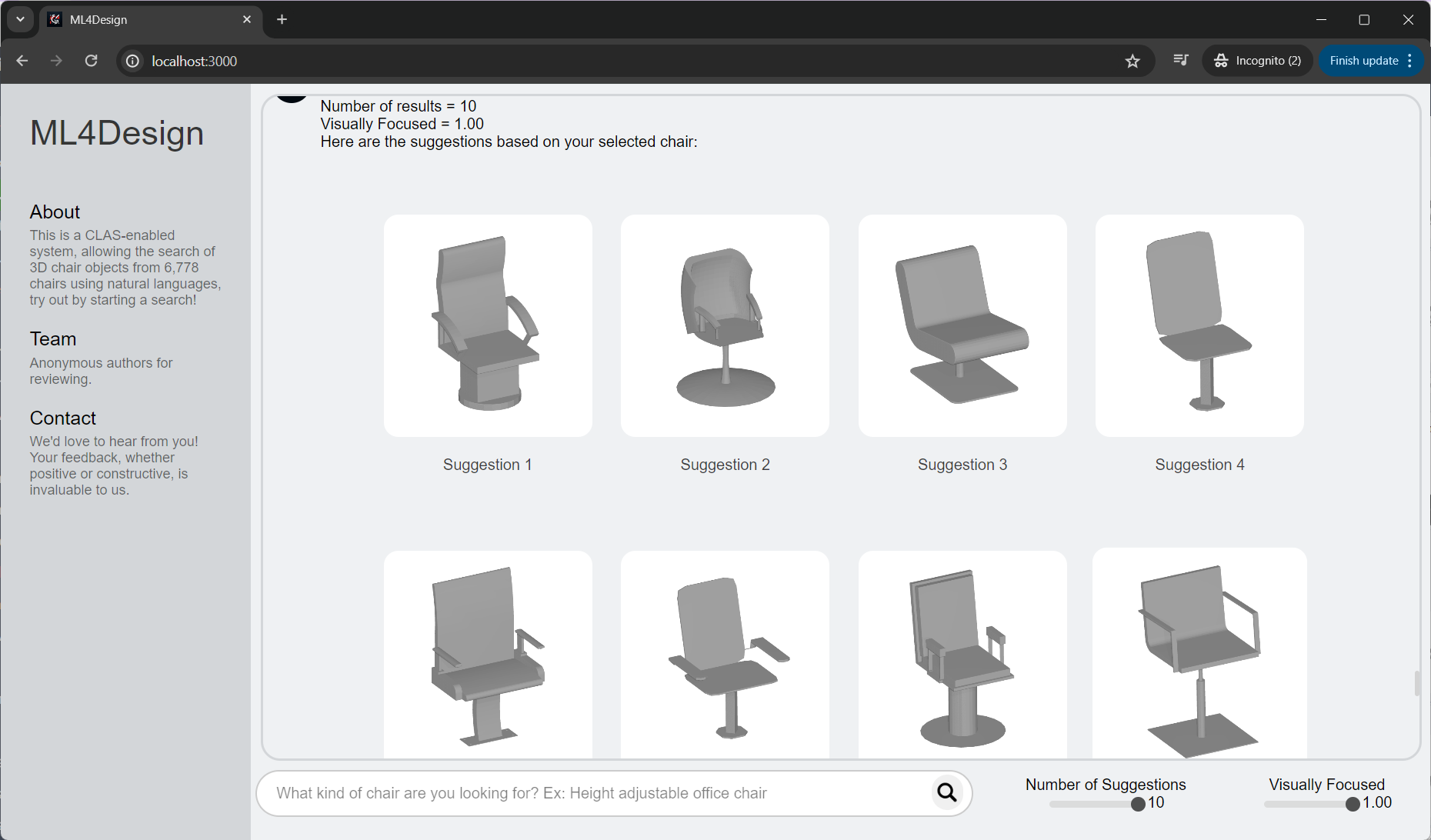}
    \caption{Exemplary use case by Peter. Search based on the selected suggestion.}
    \label{fig:story-3}
\end{figure}

Peter is a 3D model designer working at a famous office chair company. His daily job involves proposing new ideas and creating CAD designs for the company's next product. Typically, he searches online for pictures, reads books, and discusses ideas with teammates to come up with new concepts. However, after years of work, he has started to struggle with generating innovative ideas that can satisfy his manager. In searching for inspirations, a search system enabled by \clas{} can be helpful.

This time, his design challenge is to propose a modern minimalist office chair suitable for reading novels, targeting a specific market. He typed ``\textit{modern minimalist office chair suitable for reading novels}" into the system and chose to receive 10 suggestions to maximize the variety of ideas as shown in Figure~\ref{fig:story-1}. The system quickly retrieved 10 models within seconds, providing precise 3D designs. Peter was impressed by the third suggestion and clicked on it for more details, as shown in Figure~\ref{fig:story-2}. After reading the description and confirming it matched his requirements, he used the ``\textit{search similar}" feature to find more related suggestions, as shown in Figure~\ref{fig:story-3}. This time, the second suggestion caught his eye and perfectly aligned with his design needs. He downloaded the model as a reference and worked on it to further refine the design. 

This example illustrates how \clas{}-enabled search system can assist designers during the ``\textit{ideate}" and ``\textit{prototyping}" phases. \clas{}-enabled search system not only helps designers overcome creative blocks but also reduces the time they spend on 3D modeling work.

\newpage

\clearpage
\section{Evaluate effectiveness of prompts for generating descriptions of chairs}\label{appx:prompt}

\begin{figure}[h]
  \centering
  \includegraphics[width=0.3\linewidth]{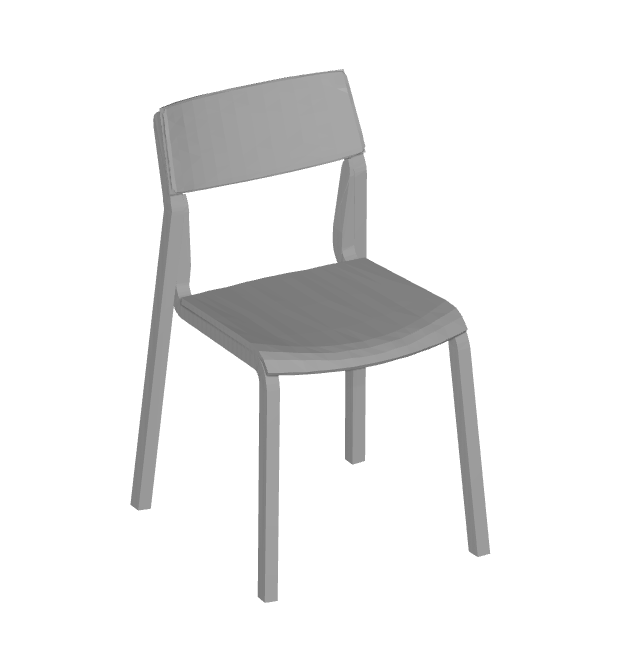}
  \caption{Chair for Testing}
  \label{fig:Chair for Testing}
\end{figure}

\begin{figure}[h!]
\centering
\scriptsize
\begin{tabular}{cccc}
\parbox[t]{0.3\textwidth}{\vspace{-50pt} Can you describe the chair’s appearance?}&
\includegraphics[width=0.2\textwidth]{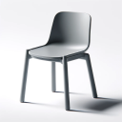} &
\includegraphics[width=0.2\textwidth]{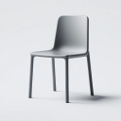} &
\includegraphics[width=0.2\textwidth]{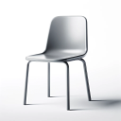} \\
\parbox{0.3\textwidth}{\vspace{-100pt} Can you describe the chair’s appearance? Try to include as many details as possible.}&
\includegraphics[width=0.2\textwidth]{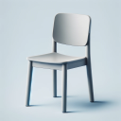} &
\includegraphics[width=0.2\textwidth]{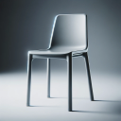} &
\includegraphics[width=0.2\textwidth]{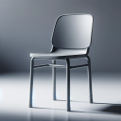} \\
\parbox{0.3\textwidth}{\vspace{-70pt} Please describe the provided object within 5 sentences. The texture, material, shadowing and color are not important. The first sentence should focus on its intended purpose and the type of user who would be most suited for this design. For the remaining four sentences, please focus on the unique geometry and proportion.
}&
\includegraphics[width=0.2\textwidth]{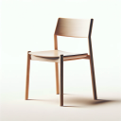} &
\includegraphics[width=0.2\textwidth]{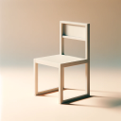} &
\includegraphics[width=0.2\textwidth]{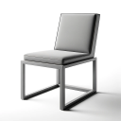} \\
\parbox{0.3\textwidth}{\vspace{-50pt} Please describe the provided object within 5 sentences. Please focus on the SHAPE, PROPORTION, and UNIQUENESS of the object.}&
\includegraphics[width=0.2\textwidth]{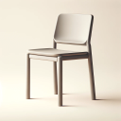} &
\includegraphics[width=0.2\textwidth]{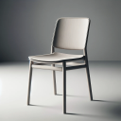} &
\includegraphics[width=0.2\textwidth]{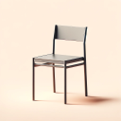} \\
\end{tabular}
\caption{Iterative process to improve structure prompt for detailed descriptions. The prompt used is at the left and the images on the right are generated by the descriptions obtained using the prompt from DALLE-E.}
\end{figure}

\clearpage
\section{CLIP architecture}\label{appx:clip}
\footnotesize
\begin{verbatim}
CLIPModel(
  (text_model): CLIPTextTransformer(
    (embeddings): CLIPTextEmbeddings(
      (token_embedding): Embedding(49408, 512)
      (position_embedding): Embedding(77, 512)
    )
    (encoder): CLIPEncoder(
      (layers): ModuleList(
        (0-11): 12 x CLIPEncoderLayer(
          (self_attn): CLIPAttention(
            (k_proj): Linear(in_features=512, out_features=512, bias=True)
            (v_proj): Linear(in_features=512, out_features=512, bias=True)
            (q_proj): Linear(in_features=512, out_features=512, bias=True)
            (out_proj): Linear(in_features=512, out_features=512, bias=True)
          )
          (layer_norm1): LayerNorm((512,), eps=1e-05, elementwise_affine=True)
          (mlp): CLIPMLP(
            (activation_fn): QuickGELUActivation()
            (fc1): Linear(in_features=512, out_features=2048, bias=True)
            (fc2): Linear(in_features=2048, out_features=512, bias=True)
          )
          (layer_norm2): LayerNorm((512,), eps=1e-05, elementwise_affine=True)
        )
      )
    )
    (final_layer_norm): LayerNorm((512,), eps=1e-05, elementwise_affine=True)
  )
  (vision_model): CLIPVisionTransformer(
    (embeddings): CLIPVisionEmbeddings(
      (patch_embedding): Conv2d(3, 768, kernel_size=(32, 32), stride=(32, 32), bias=False)
      (position_embedding): Embedding(50, 768)
    )
    (pre_layrnorm): LayerNorm((768,), eps=1e-05, elementwise_affine=True)
    (encoder): CLIPEncoder(
      (layers): ModuleList(
        (0-11): 12 x CLIPEncoderLayer(
          (self_attn): CLIPAttention(
            (k_proj): Linear(in_features=768, out_features=768, bias=True)
            (v_proj): Linear(in_features=768, out_features=768, bias=True)
            (q_proj): Linear(in_features=768, out_features=768, bias=True)
            (out_proj): Linear(in_features=768, out_features=768, bias=True)
          )
          (layer_norm1): LayerNorm((768,), eps=1e-05, elementwise_affine=True)
          (mlp): CLIPMLP(
            (activation_fn): QuickGELUActivation()
            (fc1): Linear(in_features=768, out_features=3072, bias=True)
            (fc2): Linear(in_features=3072, out_features=768, bias=True)
          )
          (layer_norm2): LayerNorm((768,), eps=1e-05, elementwise_affine=True)
        )
      )
    )
    (post_layernorm): LayerNorm((768,), eps=1e-05, elementwise_affine=True)
  )
  (visual_projection): Linear(in_features=768, out_features=512, bias=False)
  (text_projection): Linear(in_features=512, out_features=512, bias=False)
)

\end{verbatim}

\clearpage
\section{Retrieval results}\label{appx:results}
\begin{figure}[h]
    \centering
    \includegraphics[width=0.8\linewidth]{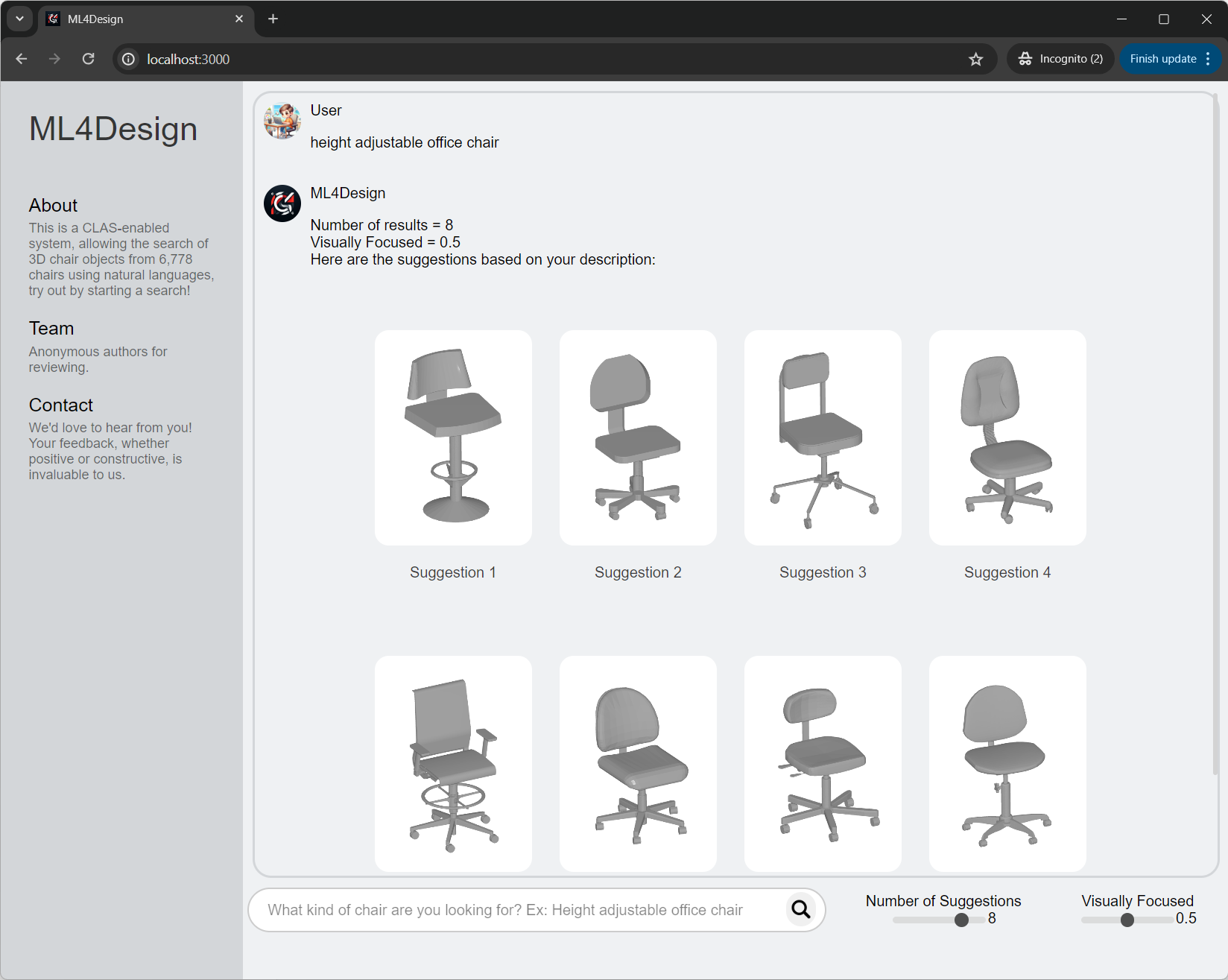}
    \caption{Height adjustable office chair.}
\end{figure}

\begin{figure}[h]
    \centering
    \includegraphics[width=0.8\linewidth]{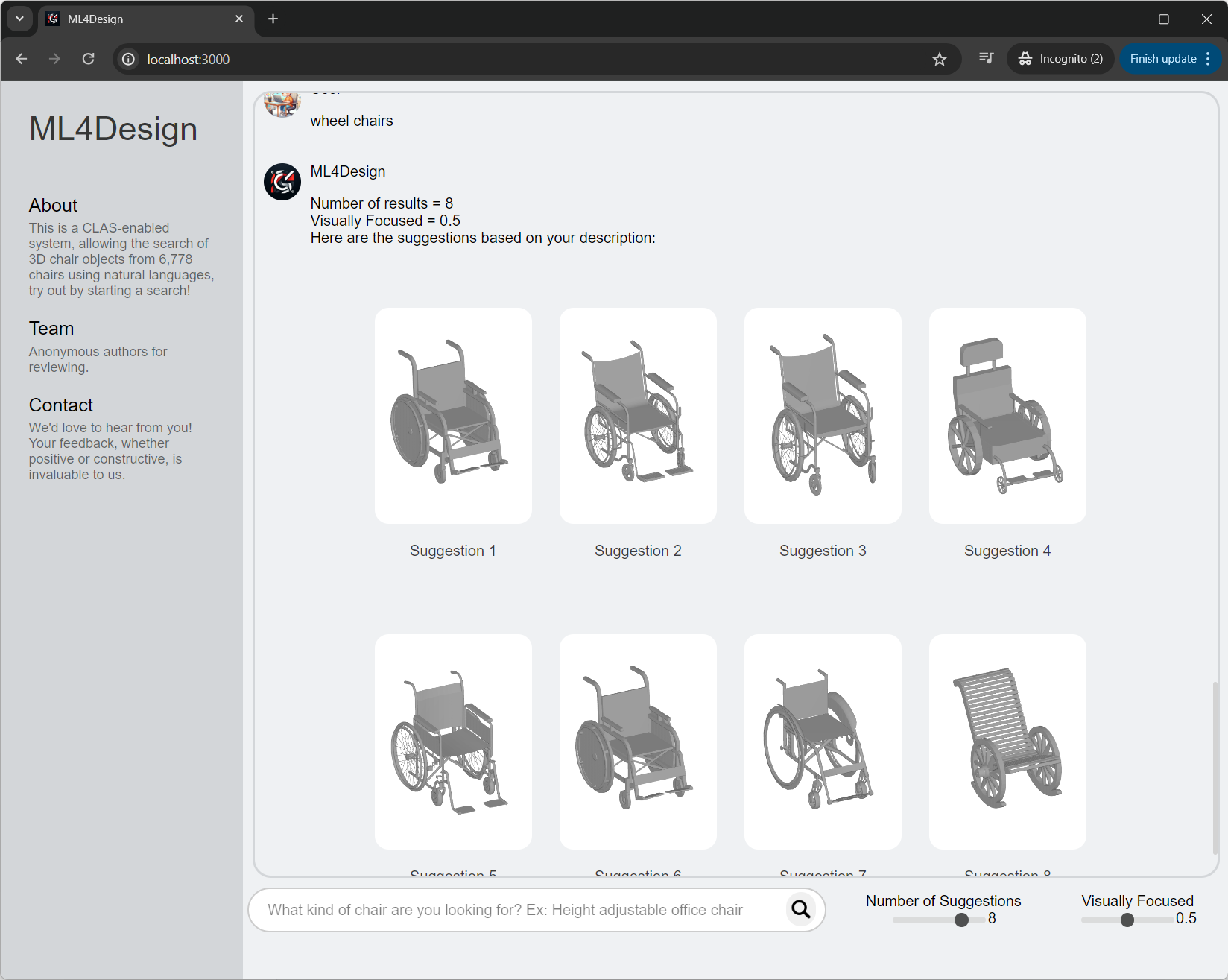}
    \caption{Wheel chairs.}
\end{figure}

\begin{figure}[h]
    \centering
    \includegraphics[width=0.85\linewidth]{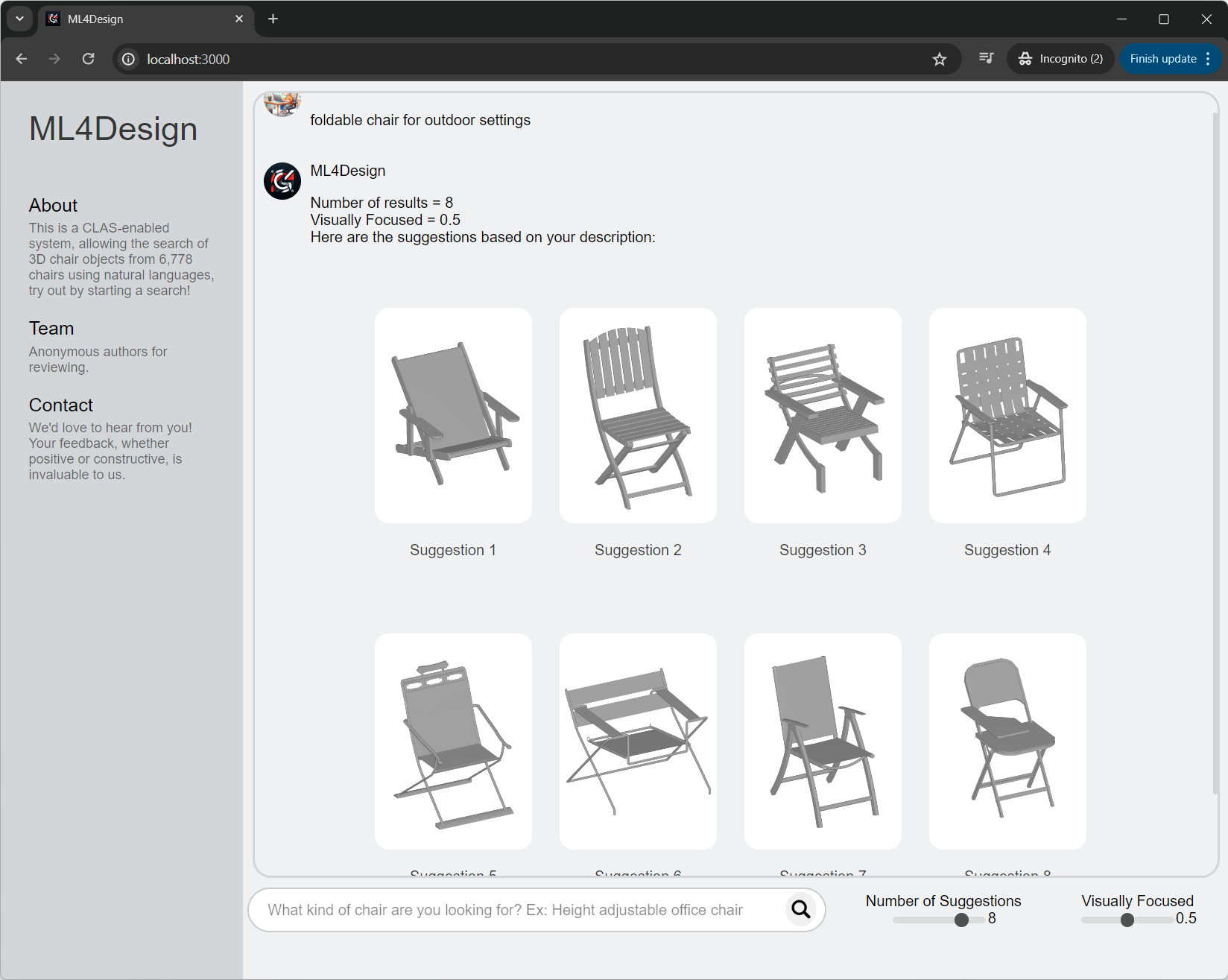}
    \caption{Foldable chair for outdoor settings.}
\end{figure}

\begin{figure}[h]
    \centering
    \includegraphics[width=0.85\linewidth]{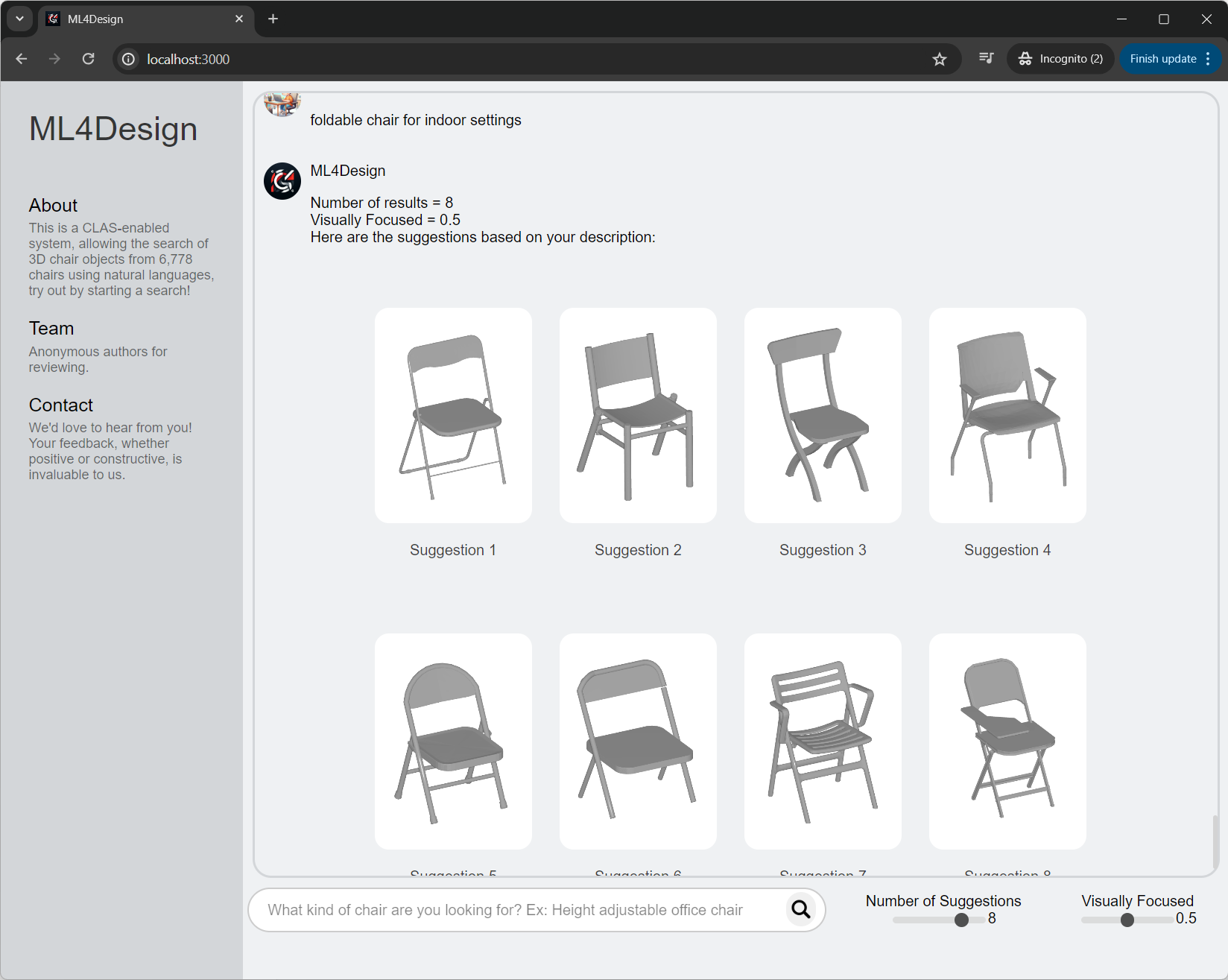}
    \caption{Foldable chair for indoor settings.}
\end{figure}

\begin{figure}[h]
    \centering
    \includegraphics[width=0.85\linewidth]{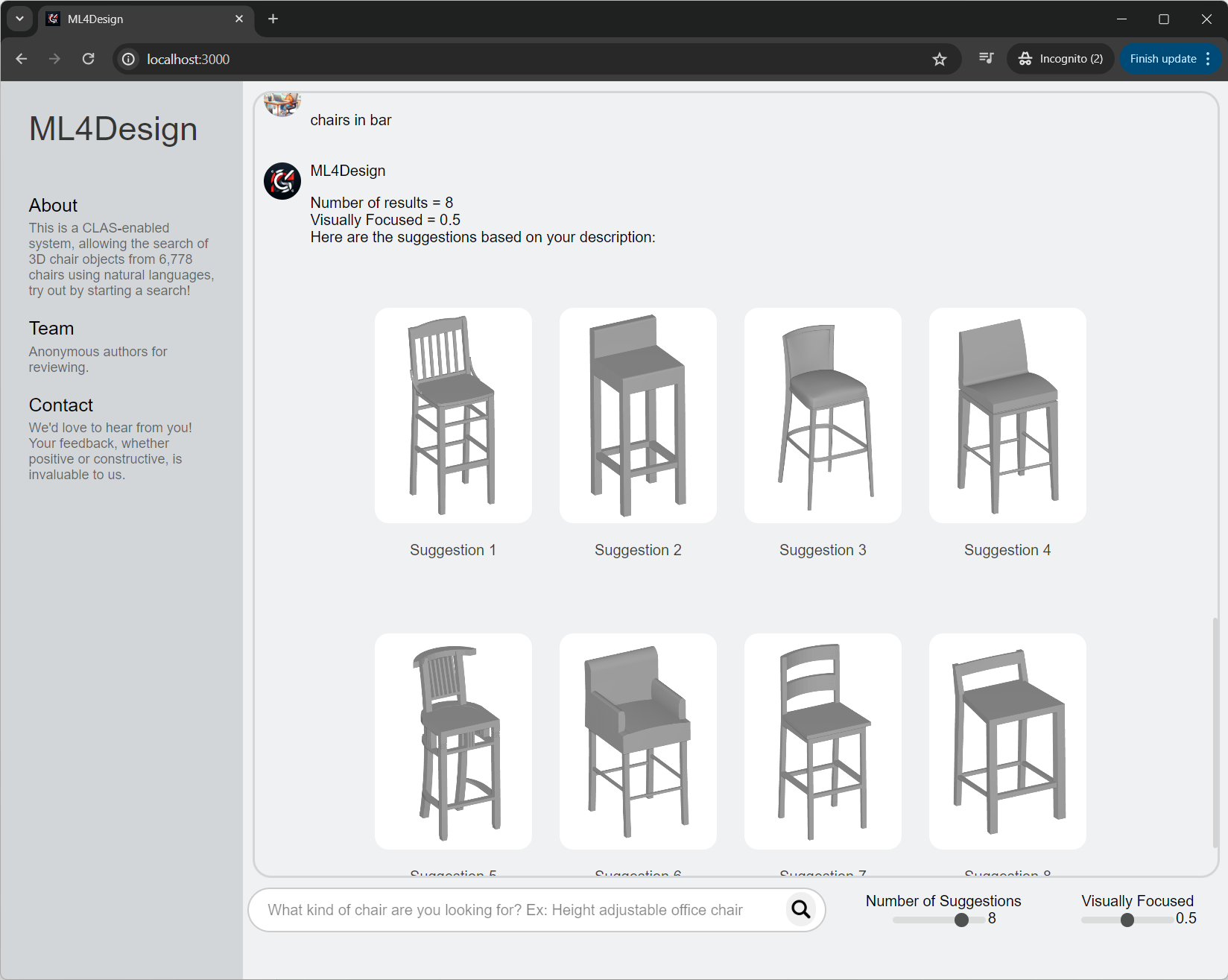}
    \caption{Chair in bar.}
\end{figure}

\begin{figure}[h]
    \centering
    \includegraphics[width=0.85\linewidth]{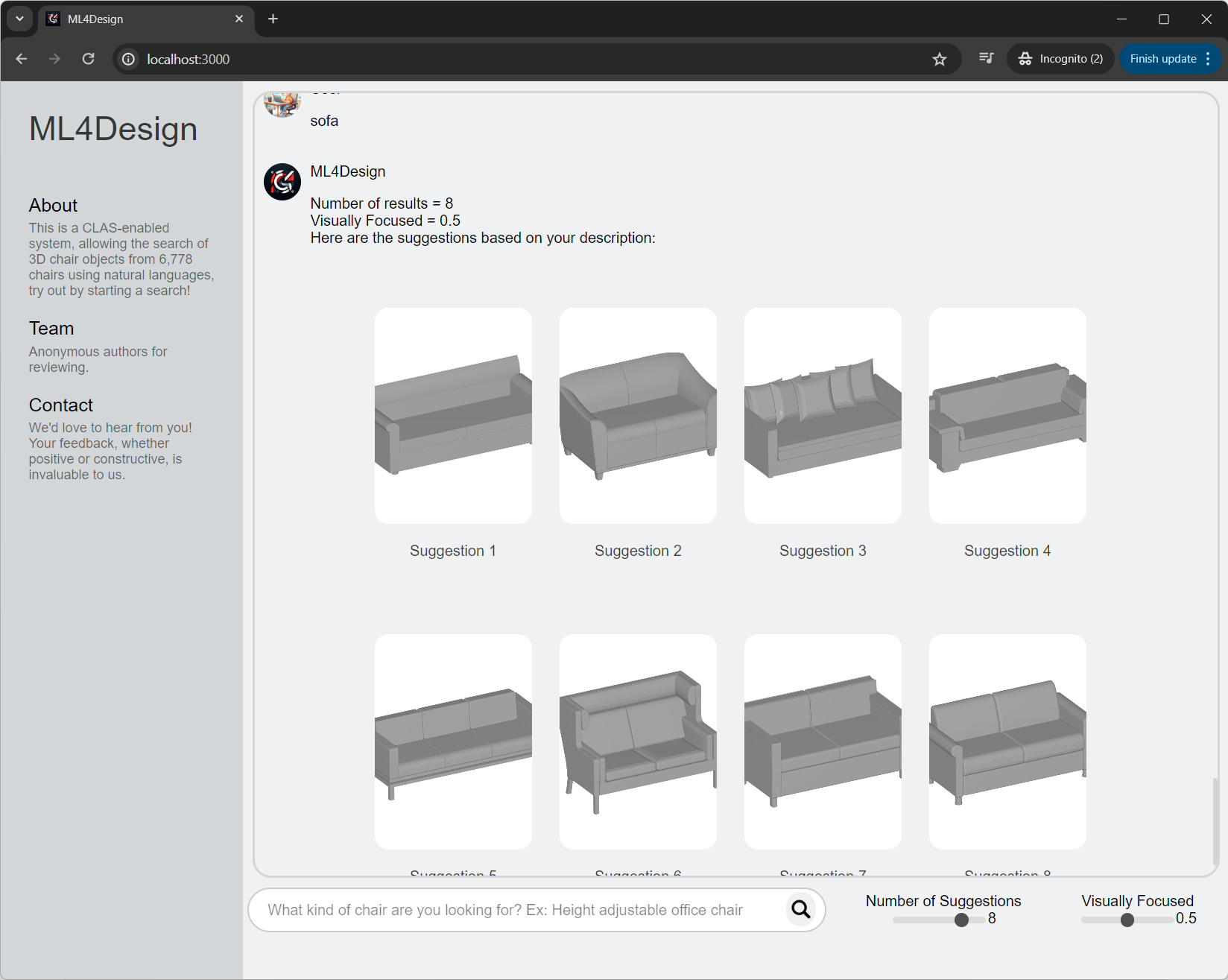}
    \caption{Sofa.}
\end{figure}

\end{document}